\documentclass[letterpaper]{article} % DO NOT CHANGE THIS

\ifdefined\aaaianonymous
    \usepackage[submission]{main}  % Anonymous submission version
\else
    \usepackage{main}              % Camera-ready version
\fi

\usepackage{times}  % DO NOT CHANGE THIS
\usepackage{helvet}  % DO NOT CHANGE THIS
\usepackage{courier}  % DO NOT CHANGE THIS
\usepackage[hyphens]{url}  % DO NOT CHANGE THIS
\usepackage{graphicx} % DO NOT CHANGE THIS
\usepackage{natbib}  % DO NOT CHANGE THIS AND DO NOT ADD ANY OPTIONS TO IT
\usepackage{caption} % DO NOT CHANGE THIS AND DO NOT ADD ANY OPTIONS TO IT
\usepackage{algorithm}
\usepackage{algorithmic}

\usepackage{newfloat}
\usepackage{listings}
\DeclareCaptionStyle{ruled}{labelfont=normalfont,labelsep=colon,strut=off} % DO NOT CHANGE THIS
\floatstyle{ruled}
\newfloat{listing}{tb}{lst}{}
\floatname{listing}{Listing}

\usepackage{amsmath} 

\usepackage{tabularray}
\usepackage{tabularx} 
\usepackage{multirow}
\usepackage{booktabs}
\usepackage{makecell}
\usepackage{array}  
\usepackage{graphicx} % Required for inserting images
\usepackage{pgfplots}
\usepackage{pgfplotstable}
\usepackage{tikz} % Required for patterns and drawing arrows
\usetikzlibrary{patterns} % For hatched patterns
\pgfplotsset{compat=1.18}
\usetikzlibrary{patterns}
\usepackage{xcolor}
\definecolor{myCoral}{HTML}{E99C93}       
\definecolor{myLavenderBlue}{HTML}{9FACD3} 
\definecolor{myPeach}{HTML}{F1DBB9}        
\definecolor{myLilac}{HTML}{D9D1E3}        
\definecolor{myPeriwinkle}{HTML}{CAD4E7}   
\usepackage{subcaption}

\usepackage{hyperref}
\hypersetup{
    hidelinks,
    colorlinks=true,
    linkcolor=blue,
    citecolor=blue,
    urlcolor=blue
}
\nocopyright

\title{Why Do MLLMs Struggle with Spatial Understanding? \\ A Systematic Analysis from Data to Architecture}

\author{
    Wanyue Zhang\textsuperscript{\rm 1,2},
    Yibin Huang\textsuperscript{\rm 4},
    Yangbin Xu\textsuperscript{\rm 6},
    JingJing Huang\textsuperscript{\rm 3}, \\
    Helu Zhi\textsuperscript{\rm 4}, 
    Shuo Ren\textsuperscript{\rm 1}, 
    Wang Xu\textsuperscript{\rm 3}\thanks{Co-corresponding author.}, 
    Jiajun Zhang \textsuperscript{\rm 1,2,5}\footnotemark[1]
}
\affiliations{
    \textsuperscript{\rm 1}Institute of Automation, Chinese Academy of Sciences\\
    \textsuperscript{\rm 2}School of Artificial Intelligence, University of Chinese Academy of Sciences\\
    \textsuperscript{\rm 3}Tsinghua University ~\textsuperscript{\rm 4}Harbin Institute of Technology ~\textsuperscript{\rm 5}Wuhan AI Research\\
    \textsuperscript{\rm 6}Institute of Microelectronics of the Chinese Academy of Sciences\\
    zhangwanyue2023@ia.ac.cn, shuo.ren@ia.ac.cn, xwjim812@gmail.com, jjzhang@nlpr.ia.ac.cn
}

\usepackage{bibentry}
\begin{document}

\maketitle

\begin{abstract}
Spatial understanding is essential for \textbf{M}ultimodal \textbf{L}arge \textbf{L}anguage \textbf{M}odels (MLLMs) to support perception, reasoning, and planning in embodied environments. 
Despite recent progress, existing studies reveal that MLLMs still struggle with spatial understanding.
However, existing research lacks a comprehensive and systematic evaluation of these limitations, often restricted to isolated scenarios, such as single-view or video. 
In this work, we present a systematic analysis of spatial understanding from both data and architectural perspectives across three representative scenarios: single-view, multi-view, and video.
We propose a benchmark named MulSeT (\textbf{Mul}ti-view \textbf{S}patial Und\textbf{e}rstanding \textbf{T}asks)\footnote{The benchmark MulSeT is available at \url{https://huggingface.co/datasets/WanyueZhang/MulSeT}.}, and design a series of experiments to analyze the spatial reasoning capabilities of MLLMs\footnote{The code is available at \url{https://github.com/WanyueZhang-ai/spatial-understanding}}.
From the data perspective, the performance of spatial understanding converges quickly as the training data increases, and the upper bound is relatively low, especially for tasks that require spatial imagination.
This indicates that merely expanding training data is insufficient to achieve satisfactory performance.
From the architectural perspective, we find that spatial understanding relies more heavily on the positional encoding within the visual encoder than within the language model, in both cascaded and native MLLMs.
Moreover, we explore reasoning injection and envision future improvements through architectural design to optimize spatial understanding. 
These insights shed light on the limitations of current MLLMs and suggest new directions for improving spatial reasoning capabilities through data scaling and architectural tuning.
\end{abstract}

\section{Introduction}

%====================================================================
\begin{figure}[t]
  \centering
  \begin{tikzpicture}
    \begin{axis}[
      hide axis,
      xmin=0, xmax=1,
      ymin=0, ymax=1,
      width=5cm,
      height=3cm,
      legend style={
        at={(0.5,1.05)},
        anchor=south,
        legend columns=3,
        font=\scriptsize,
        /tikz/every even column/.append style={column sep=0.3cm},
        draw=none,
        fill=none
      },
      legend image code/.code={
        \draw[#1,fill=#1] (0cm,-0.1cm) rectangle (0.25cm,0.08cm);
      }
    ]
      \addlegendimage{fill=myPeach!30}
      \addlegendimage{fill=myLilac}
      \addlegendimage{fill=myPeriwinkle}
      \legend{{Human-level}, {LLaVA-OV-7B}, {Qwen2.5-VL-7B}}
    \end{axis}
  \end{tikzpicture}
  
  \begin{minipage}{0.48\columnwidth} 
  \centering
  \begin{tikzpicture}
    \begin{axis}[
        name=main,
        width=\columnwidth,
        height=3cm,
        ybar=2pt,
        axis lines*=left,
        ymin=0, ymax=1.1,
        xmin=-0.5, xmax=2.5,
        xtick={0,1,2},
        xticklabels={What'sUp, COCO, VG},
        xticklabel style={
            font=\scriptsize, 
            yshift=-5pt,
            rotate=25,      % 斜向上旋转
            anchor = center,},
        ylabel style={font=\scriptsize},
        yticklabel style={font=\scriptsize},
        ymajorgrids=true,
        grid style=dashed,
        bar width=5pt,
    ]
    \addplot[bar shift=0, fill=myPeach!30] coordinates {(-0.24,1.0) (0.76,0.973) (1.64, 0.99)}; 

    \addplot[bar shift=0, fill=myPeriwinkle] coordinates {(0.24,  0.984) (1.24, 0.865) (2.12, 0.944)}; 
    
    \addplot[bar shift=0, fill=myLilac] coordinates {(0.0, 0.867) (1.0, 0.874) (1.88, 0.906)}; 
    \end{axis}
  
    \end{tikzpicture}
    \subcaption{Single-View}
    \label{align_1}
  \end{minipage}
  \hspace{0.01\columnwidth}
  \begin{minipage}{0.48\columnwidth} 
  \centering
  \begin{tikzpicture}
    \begin{axis}[
        name=main,
        width=\columnwidth,
        height=3cm,
        ybar=2pt,
        axis lines*=left,
        ymin=0, ymax=1.1,
        xmin=-0.5, xmax=2.5,
        xtick={0,1,2},
        xticklabels={
          Dis. Com.,
          Occ. Res.,
          Azi. Tra.,
        },
        xticklabel style={
           font=\scriptsize, 
           yshift=-5pt,
           rotate=25,      % 斜向上旋转
           anchor = center,},     % 对齐方式，配合旋转方向
        ylabel style={font=\scriptsize},
        yticklabel style={font=\scriptsize},
        ymajorgrids=true,
        grid style=dashed,
        bar width=5pt,
    ]
    \addplot[bar shift=0, fill=myPeach!30] coordinates {(-0.24,1.0) (0.76,0.973) (1.64, 0.99)}; 

    \addplot[bar shift=0, fill=myPeriwinkle] coordinates {(0.24,  0.4375) (1.24, 0.2239) (2.12, 0.2468)}; 
    
    \addplot[bar shift=0, fill=myLilac] coordinates {(0.0, 0.4341) (1.0, 0.3452) (1.88, 0.2725)}; 
    \end{axis}
  
    \end{tikzpicture}
    \subcaption{Multi-View}
  \end{minipage}
  \hfill

  \begin{minipage}{\columnwidth} 
  \centering
  \begin{tikzpicture}
    \begin{axis}[
        name=main,
        width=\columnwidth,
        height=3cm,
        ybar=2pt,
        axis lines*=left,
        ymin=0, ymax=1.1,
        xmin=-0.5, xmax=5.5,
        xtick={0,1,2,3,4,5},
        xticklabels={
          Abs. Dist.,
          Route Plan,
          Rel. Dir.,
          Obj. Size,
          Rel. Dist.,
          Appr. Order,
          Room Size,
          Obj. Count},
        xticklabel style={
          font=\scriptsize, 
          yshift=-5pt,
          rotate=25,      % 斜向上旋转
          anchor = center,     % 对齐方式，配合旋转方向
        },
        ylabel style={font=\scriptsize},
        yticklabel style={font=\scriptsize},
        ymajorgrids=true,
        grid style=dashed,
        bar width=5pt,
    ]
    \addplot[bar shift=0, fill=myPeach!30] coordinates {(-0.24,0.47) (0.76,0.958) (1.76, 0.958) (2.76,0.604) (3.76,0.947) (4.76,1.0)}; 

    \addplot[bar shift=0, fill=myLilac] coordinates {(0.0, 0.202) (1.0, 0.294) (2.0, 0.352) (3.0, 0.474) (4.0, 0.474) (5.0,0.244)}; 
    
    \addplot[bar shift=0, fill=myPeriwinkle] coordinates {(0.24, 0.212) (1.24, 0.3) (2.24, 0.395) (3.24, 0.493) (4.24, 0.373) (5.24,0.296)}; 
    
    \end{axis}
  
    \end{tikzpicture}
    \subcaption{Video}
  \end{minipage}
  
\caption{Comparison of MLLMs and human performance on spatial understanding benchmarks. } 
\label{fig:computation_confirm}
\end{figure}
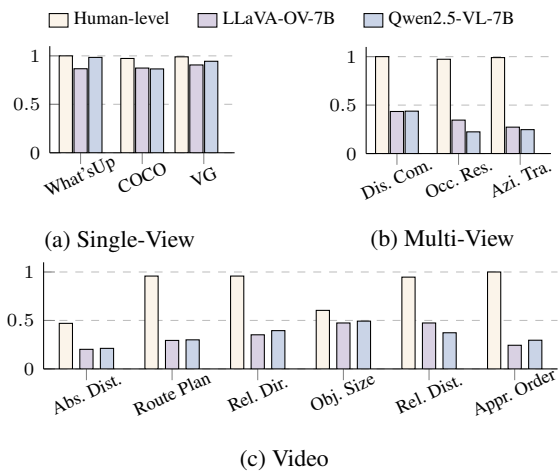
%====================================================================

Spatial understanding~\cite{yang2025thinkingspacemultimodallarge} is a core capability for MLLMs~\cite{hurst2024gpto, zhu2025internvl3exploringadvancedtraining} to support perception, reasoning, and planning in embodied scenarios~\cite{cheng2025embodiedevalevaluatemultimodalllms, du-etal-2024-embspatial}. 
It plays a central role in cognitive tasks such as object manipulation~\cite{li2023manipllmembodiedmultimodallarge, xiong2024aicmllmautonomousinteractive}, path planning~\cite{colan2025assessingvaluevisualinput}, and developing world models~\cite{liu2024world}. 
Spatial understanding requires integrating multi-modal information, modeling geometric relationships, and maintaining spatial consistency across perspectives. 
As shown in Figure~\ref{fig:computation_confirm}, while MLLMs achieve human-level performance on single-view spatial tasks, a significant gap emerges in multi-view and video-based scenarios, revealing a fundamental deficit in their spatial understanding capabilities.

Recent studies have made efforts to analyze the potential mechanism~\cite{yang2025thinkingspacemultimodallarge,chen2025spatialreasoninghardvlms}. 
\citet{chen2025spatialreasoninghardvlms} examine spatial reasoning from an attention mechanism perspective, identifying a strong link between accurate attention alignment and success on spatial tasks. But such insights are often derived from narrow task settings—typically involving single-view direction determination.
\citet{yang2025thinkingspacemultimodallarge} reveal that generating cognitive maps during question-answering enhances MLLMs' spatial distance ability.
However, despite these insights, existing analyses fall short in capturing the complexity of spatial understanding, motivating a more comprehensive investigation. 

%====================================================================
\begin{figure*}[t!]
\centering
\includegraphics[width=1\textwidth]{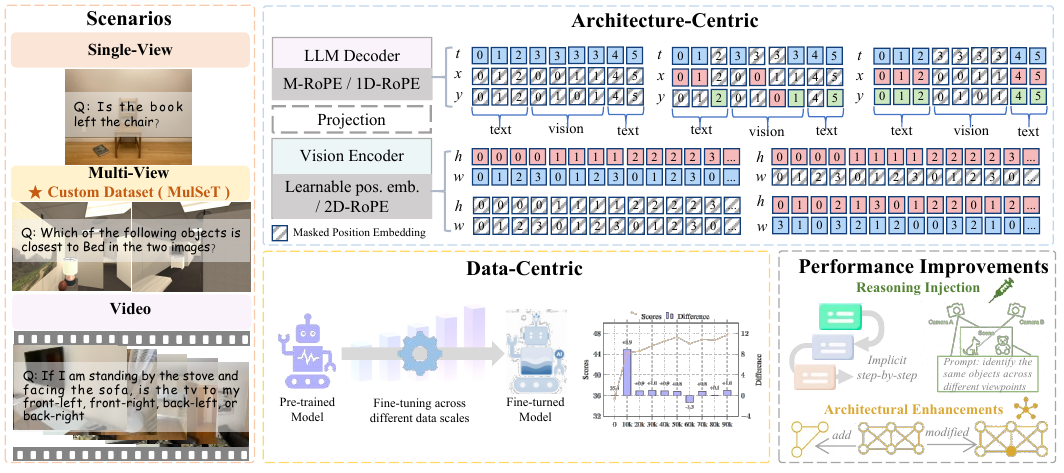}
\caption{Overview of our study on spatial reasoning in MLLMs. The terms t, x, y, h, and w denote different dimensions of position embeddings.}
\label{fig:overview}\vspace{-6mm} 
\end{figure*}
%====================================================================

To address this issue, we analyze the underlying reasons for the poor spatial reasoning ability systematically. 
This leads us to pose a central research question: Are current MLLMs limited in spatial understanding primarily due to insufficient training data, or are they fundamentally constrained by their architectural design?
To investigate this question, we conduct abundant experiments covering single-view, multi-view, and videos.
For multiple images, we propose MulSeT benchmark, systematically evaluating MLLMs' spatial reasoning capabilities across three representative subtasks: occlusion restoration, distance comparison, and azimuth transfer. 
MulSeT features three representative subtasks meticulously designed to probe a gradient of spatial understanding. 
Specifically, occlusion restoration needs to understand the relative positions between views, distance comparison requires intuitive spatial understanding, and azimuth transfer demands spatial imagination.

From the data perspective, our analysis of training data scaling reveals that performance gains are not uniform across tasks, but rather correlate strongly with our proposed gradient from spatial matching to abstract reasoning. On tasks could be solved by spatial matching related to semantic, we observe that MLLM performance improves substantially and consistently with more data, eventually plateauing at a high level. Conversely, as tasks shift towards abstract reasoning, the upper bound of performance drops significantly. These observations suggest that increasing training data alone may offer diminishing returns for spatial understanding, and that data scale, in isolation, may not be sufficient to drive meaningful improvements across tasks.

From the architectural perspective, we conduct controlled ablation studies on positional encoding mechanisms in both the visual and language components. We find that spatial understanding relies predominantly on the visual encoder’s positional encodings, with limited contributions from the language model—especially in single-view tasks. 
These findings suggest a structural bottleneck in the spatial reasoning pathway of current MLLMs.

Moreover, we explore and discuss methods to optimize the spatial understanding ability, including reasoning injection and the model architecture.
We propose a multi-view prompting strategy to improve the performance.

In summary, our key contributions are as follows:

\begin{itemize}
    \item We propose and release a multi-perspective benchmark, MulSeT, for spatial understanding, encompassing three challenging spatial reasoning subtasks.
    
    \item Through empirical analysis, we reveal that merely scaling up training data is insufficient to significantly improve spatial understanding, especially for tasks needing spatial imagination.
    Moreover, positional encoding in the visual encoder is a crucial factor influencing spatial capability.
    
    \item We explore reasoning injection as a potential enhancement strategy, and suggest that future spatial reasoning improvements require structural model design beyond data augmentation. 
     Furthermore, we propose a multi-view prompting strategy, and our experiments demonstrate its effectiveness.
\end{itemize}

\begin{table}[t!]
\centering
\small
\caption{Human evaluation results on MulSeT.}
\begin{tabular}{lccc}
\toprule
\textbf{Evaluator} 
& \makecell{\textbf{Distance} \\ \textbf{Comparison}} 
& \makecell{\textbf{Occlusion} \\ \textbf{Restoration}} 
& \makecell{\textbf{Azimuth} \\ \textbf{Transfer}}  \\
\midrule
\textbf{Average} & 86.25\% & 57.50\% & 51.25\% \\
\bottomrule
\end{tabular}
\label{tab:human_eval}
\end{table}

\section{MulSeT Benchmark}
\label{sec:mulset_benchmark}

%====================================================================
\begin{figure*}[t!]
\centering
\includegraphics[width=1\textwidth]{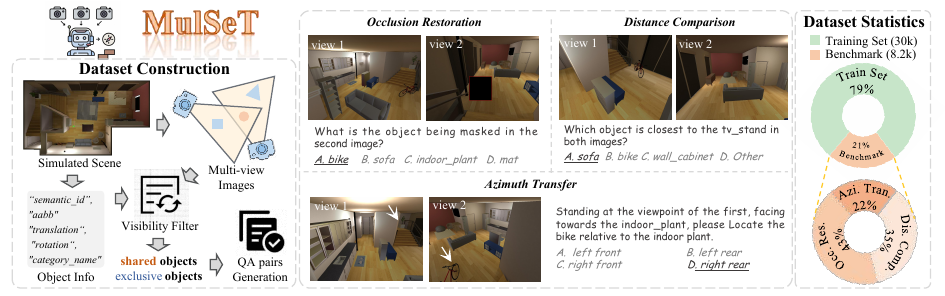}
 \caption{Overview of MulSeT.}\vspace{-6mm}
\label{fig:dataset}
\end{figure*}
%====================================================================

To systematically analyze the spatial understanding capabilities of MLLMs, we evaluate MLLMs from three representative visual scenarios: single-view, multi-view, and video-based tasks.
As for benchmark of multiple images, there is currently a lack of high-quality benchmarks tailored for multi-view spatial understanding tasks~\cite{yeh2025seeingperspectiveevaluatingmultiview, jia2025omnispatialcomprehensivespatialreasoning}, which are crucial for assessing models' ability to integrate spatial cues across different viewpoints or complementary images. To fill this gap, we introduce \textbf{MulSeT}, a simulation-based dataset designed to systematically evaluate MLLMs' spatial understanding in multi-view settings.
MulSeT enables controlled comparisons and targeted diagnosis of model behavior, serving as a necessary complement to existing single-view and video-based datasets.

\subsection{Task Design}

To evaluate multi-view spatial understanding, we designed three progressively challenging tasks that require integrating information across images.

\textbf{Occlusion Restoration} requires understanding the relative positions between views.
Given two views of a scene, one object in the second image is masked. The model must identify the occluded object by leveraging information from both views, testing object correspondence in different views.

\textbf{Distance Comparison} requires intuitive spatial understanding.
The model is asked to find the object closest to a given reference object (shared across views) based on centroid distance, assessing spatial relation inference.

\textbf{Azimuth Transfer} requires abstract spatial imagination and viewpoint-conditioned spatial reasoning.
Assuming an egocentric viewpoint from the first image while facing a reference object, the model must determine the relative direction of a second object in the second view.

\subsection{Construction Pipeline}

Thanks to the controllability of the simulator, we construct the dataset in a fully synthetic environment. Multiple indoor scenes are built in the simulator, where two images are captured per scene from different viewing angles. Each image pair contains both shared and exclusive objects, ensuring that some objects appear in only one of the two views.

To construct each sample, we first obtain object-level scene metadata, including the 3D positions of objects in the world coordinate system. Based on this, we identify shared and exclusive objects across views. An automated script then traverses all object candidates, applying a visibility filter that retains only objects occupying at least a minimum area ratio (\texttt{min\_area\_ratio}) in the image. Following this, we generate the specific tasks. For \textbf{Occlusion Restoration}, we select a shared object and mask it in one view with a black rectangle bordered in red. For \textbf{Distance Comparison}, we choose a shared object and use the 3D coordinates of all objects to generate a question-answer pair about which is closest. For \textbf{Azimuth Transfer}, we select two exclusive objects and use their coordinates to compute the relative azimuth, enforcing an angular separation constraint (\texttt{angle\_thresh\_deg} $\geq$ 15$^\circ$) to ensure sufficient directional distinction.

The tasks are designed with increasing difficulty, and all are formulated as four-option multiple-choice questions. Examples are shown in Figure~\ref{fig:dataset}. Human evaluation results are shown in Table~\ref{tab:human_eval}

Our benchmark contains over 38.2k question-answer pairs spanning more than 5,000 unique 3D scenes, with source imagery from the validation sets of AI2THOR~\cite{kolve2017ai2}. %ScanNet 
For the following experiments, we use 30k data for training and 8.2k data for evaluation.

\section{Data-Centric Analysis}

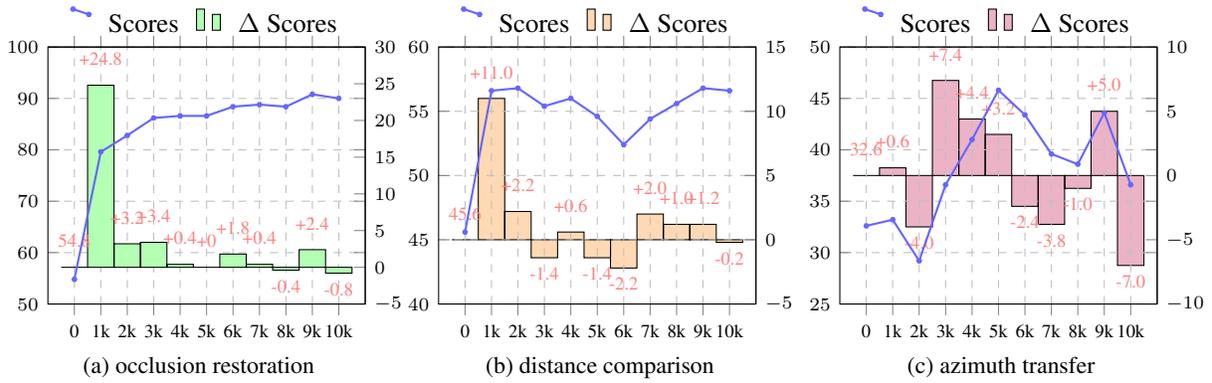
\begin{figure*}[t!]
  \centering
  \begin{subfigure}{0.3\linewidth}
    \centering
    \pgfplotsset{set layers}
    \begin{tikzpicture}

        \pgfplotscreateplotcyclelist{mycolors}{
            {fill=red!30},% baseline
            {fill=blue!30},% 1k
            {fill=green!30},% 2k
            {fill=orange!30},% 3k
            {fill=purple!30},% 5k
            {fill=yellow!30},% 10k
            {fill=cyan!30},% 30k
            {fill=magenta!30},% 50k
            {fill=lime!30},% 70k
            {fill=teal!30}% 90k
        }
        
        \begin{axis}[
            on layer={axis background},
            axis y line*=right,
            ymin=-5,
            ymax=30,
            ylabel style={yshift=5pt},
            ytick={-5,0,5,10,15,20,25,30},
            yticklabel style={font=\scriptsize}, 
            grid=none, 
            enlarge x limits=0.1,
            symbolic x coords={0,1k,2k,3k,4k,5k,6k,7k,8k,9k,10k},
            xtick=\empty, 
            width=5.8cm,
            height=5cm,
            cycle list name=mycolors,
            nodes near coords, % Simply enable nodes near coords
            point meta=explicit symbolic,
            every node near coord/.append style={
                font=\scriptsize,
                yshift=2pt,
                anchor=south,
                color=red!50 %
            }
        ]

        \addplot+[
            ybar, 
            bar shift=0pt, 
            fill=green!30,
            nodes near coords,
            every node near coord/.append style={
                font=\scriptsize,
                color=red!50,
                yshift=2pt,
                anchor=south
            }
        ] coordinates {
            (0,0) [54.8]
            (1k,24.8) [+24.8]
            (2k,3.2) [+3.2]
            (3k,3.4) [+3.4]
            (4k,0.4) [+0.4]
            (5k,0) [+0]
            (6k,1.8) [+1.8]
            (7k,0.4) [+0.4]
            (9k,2.4) [+2.4]
        };

        \addplot+[
            ybar, 
            bar shift=0pt, 
            fill=green!30,
            nodes near coords,
            every node near coord/.append style={
                font=\scriptsize,
                color=red!50,
                yshift=-2pt,
                anchor=north
            }
        ] coordinates {
            (8k,-0.4) [-0.4]
            (10k,-0.8) [-0.8]
        };

        \end{axis}

        \begin{axis}[
            ybar, 
            bar width=12pt, 
            ymin=50,
            ymax=100,
            ylabel style={yshift=-5pt},
            ytick={40,50,60,70,80,90,100},
            yticklabel style={font=\scriptsize}, 
            grid=major,
            grid style={dashed, gray!50},
            enlarge x limits=0.1,
            legend style={
                at={(0.5,1)},
                anchor=south,
                column sep=1ex,
                draw=none,
                fill=none,
                legend columns=-1
            },
            symbolic x coords={0,1k,2k,3k,4k,5k,6k,7k,8k,9k,10k},
            xtick={0,1k,2k,3k,4k,5k,6k,7k,8k,9k,10k},
            xticklabel style={
                rotate=0, 
                anchor=north,
                font=\scriptsize
            },
            width=5.8cm,
            height=5cm,
            cycle list name=mycolors,
            nodes near coords, % Simply enable nodes near coords
            point meta=explicit symbolic,
            every node near coord/.append style={
                font=\scriptsize,
                yshift=2pt,
                anchor=south,
                color=red!50 %
            }
        ]

        \addplot[
            sharp plot,
            color=blue!60,
            thick,
            mark=asterisk,
            mark options={scale=0.5, fill=blue, line width=1pt},
            nodes near coords={},
            point meta=x
        ] coordinates {
            (0,54.8)
            (1k,79.6)
            (2k,82.8)
            (3k,86.2)
            (4k,86.6)
            (5k,86.6)
            (6k,88.4)
            (7k,88.8)
            (8k,88.4)
            (9k,90.8)
            (10k,90.0)
        };
        \addlegendentry{Scores}
        \addlegendimage{ybar,fill=green!30}
        \addlegendentry{$\Delta$ Scores}
        \addplot+[ybar, bar shift=0pt] coordinates {(0,0) [54.8]}; 
        \addplot+[ybar, bar shift=0pt] coordinates {(1k,24.8) [+24.8]}; 
        \addplot+[ybar, bar shift=0pt] coordinates {(2k,3.2) [+3.2]}; 
        \addplot+[ybar, bar shift=0pt] coordinates {(3k,3.4) [+3.4]}; 
        \addplot+[ybar, bar shift=0pt] coordinates {(4k,0.4) [+0.4]}; 
        \addplot+[ybar, bar shift=0pt] coordinates {(5k,0) [+0]}; 
        \addplot+[ybar, bar shift=0pt] coordinates {(6k,1.8) [+1.8]}; 
        \addplot+[ybar, bar shift=0pt] coordinates {(7k,0.4) [+0.4]}; 
        \addplot+[ybar, bar shift=0pt] coordinates {(8k,-0.4) [-0.4]}; 
        \addplot+[ybar, bar shift=0pt] coordinates {(9k,2.4) [+2.4]}; 
        \addplot+[ybar, bar shift=0pt] coordinates {(10k,-0.8) [-0.8]}; 
        
        \end{axis}

    \end{tikzpicture}\vspace{-1mm}
    \caption{\label{task1}occlusion restoration}
    \label{fig:subfig1}
  \end{subfigure}%
  \begin{subfigure}{0.3\linewidth}
    \centering
    \begin{tikzpicture}

        \pgfplotscreateplotcyclelist{mycolors}{
            {fill=red!30},% baseline
            {fill=blue!30},% 1k
            {fill=green!30},% 2k
            {fill=orange!30},% 3k
            {fill=purple!30},% 5k
            {fill=yellow!30},% 10k
            {fill=cyan!30},% 30k
            {fill=magenta!30},% 50k
            {fill=lime!30},% 70k
            {fill=teal!30}% 90k
        }

        \begin{axis}[
            axis y line*=right,
            ymin=-5,
            ymax=15,
            ylabel style={yshift=5pt},
            yticklabel style={font=\scriptsize}, 
            ytick={-5,0,5,10,15},
            grid=none, 
            enlarge x limits=0.1,
            symbolic x coords={0,1k,2k,3k,4k,5k,6k,7k,8k,9k,10k},
            xtick=\empty, 
            width=5.8cm,
            height=5cm,
            cycle list name=mycolors,
            nodes near coords, % Simply enable nodes near coords
            point meta=explicit symbolic,
            every node near coord/.append style={
                font=\scriptsize,
                yshift=2pt,
                anchor=south,
                color=red!50 %
            }
        ]

        \addplot+[
            ybar, 
            bar shift=0pt, 
            fill=orange!30,
            nodes near coords,
            every node near coord/.append style={
                font=\scriptsize,
                color=red!50,
                yshift=2pt,
                anchor=south
            }
        ] coordinates {
            (0,0) [45.6]
            (1k,11.0) [+11.0]
            (2k,2.2) [+2.2]
            (4k,0.6) [+0.6]
            (7k,2.0) [+2.0]
            (8k,1.2) [+1.0]
            (9k,1.2) [+1.2]
        };
        
        \addplot+[
            ybar, 
            bar shift=0pt, 
            fill=orange!30,
            nodes near coords,
            every node near coord/.append style={
                font=\scriptsize,
                color=red!50,
                yshift=-2pt,
                anchor=north
            }
        ] coordinates {
            (3k,-1.4) [-1.4]
            (5k,-1.4) [-1.4]
            (6k,-2.2) [-2.2]
            (10k,-0.2) [-0.2]
        };

        \end{axis}

        \begin{axis}[
            ybar, 
            bar width=12pt,
            ymin=40,
            ymax=60,
            ylabel style={yshift=-5pt},
            yticklabel style={font=\scriptsize}, 
            ytick={40,45,50,55,60},
            grid=major,
            grid style={dashed, gray!50},
            enlarge x limits=0.1, 
            legend style={
                at={(0.5,1)},
                anchor=south,
                column sep=1ex,
                draw=none,
                fill=none,
                legend columns=-1
            },
            symbolic x coords={0,1k,2k,3k,4k,5k,6k,7k,8k,9k,10k},
            xtick={0,1k,2k,3k,4k,5k,6k,7k,8k,9k,10k},
            xticklabel style={
                rotate=0, 
                anchor=north,
                font=\scriptsize
            },
            width=5.8cm,
            height=5cm,
            cycle list name=mycolors,
            nodes near coords, % Simply enable nodes near coords
            point meta=explicit symbolic,
            every node near coord/.append style={
                font=\scriptsize,
                yshift=2pt,
                anchor=south,
                color=red!50 %
            }
        ]

        \addplot[
            sharp plot,
            color=blue!60,
            thick,
            mark=asterisk,
            mark options={scale=0.5, fill=blue, line width=1pt},
            nodes near coords={},
            point meta=x
        ] coordinates {
            (0,45.6)
            (1k,56.6)
            (2k,56.8)
            (3k,55.4)
            (4k,56.0)
            (5k,54.6)
            (6k,52.4)
            (7k,54.4)
            (8k,55.6)
            (9k,56.8)
            (10k,56.6)
        };
        \addlegendentry{Scores}
        \addlegendimage{ybar,fill=orange!30}
        \addlegendentry{$\Delta$ Scores}
        \addplot+[ybar, bar shift=0pt] coordinates {(0,0) [45.6]}; 
        \addplot+[ybar, bar shift=0pt] coordinates {(1k,11.0) [+11.0]}; 
        \addplot+[ybar, bar shift=0pt] coordinates {(2k,2.2) [+2.2]}; 
        \addplot+[ybar, bar shift=0pt] coordinates {(3k,-1.4) [-1.4]}; 
        \addplot+[ybar, bar shift=0pt] coordinates {(4k,0.6) [+0.6]}; 
        \addplot+[ybar, bar shift=0pt] coordinates {(5k,-1.4) [-1.4]}; 
        \addplot+[ybar, bar shift=0pt] coordinates {(6k,-2.2) [-2.2]}; 
        \addplot+[ybar, bar shift=0pt] coordinates {(7k,2.0) [+2.0]}; 
        \addplot+[ybar, bar shift=0pt] coordinates {(8k,1.2) [+1.0]}; 
        \addplot+[ybar, bar shift=0pt] coordinates {(9k,1.2) [+1.2]}; 
        \addplot+[ybar, bar shift=0pt] coordinates {(10k,-0.2) [-0.2]}; 

        \end{axis}

    \end{tikzpicture}\vspace{-1mm}
    \caption{\label{task2}distance comparison}
  \end{subfigure}%
  \begin{subfigure}{0.3\linewidth}
        \centering
    \begin{tikzpicture}

        \pgfplotscreateplotcyclelist{mycolors}{
            {fill=red!30},% baseline
            {fill=blue!30},% 1k
            {fill=green!30},% 2k
            {fill=orange!30},% 3k
            {fill=purple!30},% 5k
            {fill=yellow!30},% 10k
            {fill=cyan!30},% 30k
            {fill=magenta!30},% 50k
            {fill=lime!30},% 70k
            {fill=teal!30}% 90k
        }

        \begin{axis}[
            axis y line*=right,
            ymin=-10,
            ymax=10,
            ylabel style={yshift=5pt},
            yticklabel style={font=\scriptsize}, 
            ytick={-10,-5,0,5,10},
            grid=none, 
            enlarge x limits=0.1,
            symbolic x coords={0,1k,2k,3k,4k,5k,6k,7k,8k,9k,10k},
            xtick=\empty, 
            width=5.8cm,
            height=5cm,
            cycle list name=mycolors,
            nodes near coords, % Simply enable nodes near coords
            point meta=explicit symbolic,
            every node near coord/.append style={
                font=\scriptsize,
                yshift=2pt,
                anchor=south,
                color=red!50 %
            }
        ]

        \addplot+[
            ybar, 
            bar shift=0pt, 
            fill=purple!30,
            nodes near coords,
            every node near coord/.append style={
                font=\scriptsize,
                color=red!50,
                yshift=2pt,
                anchor=south
            }
        ] coordinates {
            (0,0) [32.6]
            (1k,0.6) [+0.6]
            (3k,7.4) [+7.4]
            (4k,4.4) [+4.4]
            (5k,3.2) [+3.2]
            (9k,5) [+5.0]
        };
        
        \addplot+[
            ybar, 
            bar shift=0pt, 
            fill=purple!30,
            nodes near coords,
            every node near coord/.append style={
                font=\scriptsize,
                color=red!50,
                yshift=-2pt,
                anchor=north
            }
        ] coordinates {
            (2k,-4) [-4.0]
            (6k,-2.4) [-2.4]
            (7k,-3.8) [-3.8]
            (8k,-1.0) [-1.0]
            (10k,-7.0) [-7.0]
        };

        \end{axis}

        \begin{axis}[
            ybar, 
            bar width=12pt, 
            ymin=25,
            ymax=50,
            ylabel style={yshift=-5pt},
            ytick={25,30,35,40,45,50},
            yticklabel style={font=\scriptsize}, 
            grid=major,
            grid style={dashed, gray!50},
            enlarge x limits=0.1, 
            legend style={
                at={(0.5,1)},
                anchor=south,
                column sep=1ex,
                draw=none,
                fill=none,
                legend columns=-1
            },
            symbolic x coords={0,1k,2k,3k,4k,5k,6k,7k,8k,9k,10k},
            xtick={0,1k,2k,3k,4k,5k,6k,7k,8k,9k,10k},
            xticklabel style={
                rotate=0, 
                anchor=north,
                font=\scriptsize
            },
            width=5.8cm,
            height=5cm,
            cycle list name=mycolors,
            nodes near coords, % Simply enable nodes near coords
            point meta=explicit symbolic,
            every node near coord/.append style={
                font=\scriptsize,
                yshift=2pt,
                anchor=south,
                color=red!50 %
            }
        ]

        \addplot[
            sharp plot,
            color=blue!60,
            thick,
            mark=asterisk,
            mark options={scale=0.5, fill=blue, line width=1pt},
            nodes near coords={},
            point meta=x
        ] coordinates {
            (0,32.6)
            (1k,33.2)
            (2k,29.2)
            (3k,36.6)
            (4k,41.0)
            (5k,45.8)
            (6k,43.4)
            (7k,39.6)
            (8k,38.6)
            (9k,43.6)
            (10k,36.6)
        };
        \addlegendentry{Scores}
        \addlegendimage{ybar,fill=purple!30}
        \addlegendentry{$\Delta$ Scores}
        \addplot+[ybar, bar shift=0pt] coordinates {(0,0) [32.6]}; 
        \addplot+[ybar, bar shift=0pt] coordinates {(1k,0.6) [+0.6]}; 
        \addplot+[ybar, bar shift=0pt] coordinates {(2k,-4) [-4.0]}; 
        \addplot+[ybar, bar shift=0pt] coordinates {(3k,7.4) [+7.4]}; 
        \addplot+[ybar, bar shift=0pt] coordinates {(4k,4.4) [+4.4]}; 
        \addplot+[ybar, bar shift=0pt] coordinates {(5k,3.2) [+3.2]}; 
        \addplot+[ybar, bar shift=0pt] coordinates {(6k,-2.4) [-2.4]}; 
        \addplot+[ybar, bar shift=0pt] coordinates {(7k,-3.8) [-3.8]}; 
        \addplot+[ybar, bar shift=0pt] coordinates {(8k,-1.0) [-1.0]}; 
        \addplot+[ybar, bar shift=0pt] coordinates {(9k,5) [+5.0]}; 
        \addplot+[ybar, bar shift=0pt] coordinates {(10k,-7.0) [-7.0]}; 
        
        \end{axis}
        
    \end{tikzpicture}\vspace{-1mm}
    \caption{\label{task3}azimuth transfer}
  \end{subfigure}
  \caption{Performance trends under increasing training data scales for multi-view scenarios. Here, $\Delta$ Scores denotes the change in performance compared to the preceding training data scale.}\vspace{-6mm}
  \label{fig:data_train_multiview_wholefig}
\end{figure*}

To investigate how data volume affects spatial understanding in MLLMs, we conduct fine-tuning with varying dataset sizes. As shown in Table~\ref{fig:computation_confirm}, MLLMs exhibit strong performance in single-view scenarios, so we focus on the multi-view and video scenarios.

\subsection{Experiment Setup}

\textbf{Datasets and Metrics.} For the multi-view setting, we fine-tune on subsets of MulSeT with training sizes ranging from 1k to 10k samples, and evaluate on the held-out MulSeT test set. For the video setting, we fine-tune on subsets of SR-91k~\cite{ouyang2025spacerreinforcingmllmsvideo}, a synthetic dataset generated via simulation to emulate the tasks in VSI-Bench~\cite{yang2025thinkingspacemultimodallarge}, with training sizes from 10k to 90k samples. Performance is then evaluated on the VSI-Bench, which provides over 5,000 question-answer pairs designed to assess visual-spatial intelligence over time.

\textbf{Model and Baseline.} We adopt Qwen2.5-VL-7B as the backbone model and perform supervised fine-tuning with varying amounts of spatial data. 
The unfine-tuned backbone serves as the baseline for comparison. The baseline is the original Qwen2.5-VL backbone, which has not undergone any task-specific fine-tuning on our datasets.

\textbf{Training Setting.}
We employ LoRA~\cite{hu2022lora} for parameter-efficient fine-tuning~\cite{xu2023parameter, zhou-etal-2024-empirical}, with distinct configurations for each modality. 
For the multi-view training, we configure LoRA with a rank of 8 and an alpha of 32. 
The model is trained for 10 epochs with a learning rate of 1e-5, using a cosine learning rate schedule with a warmup ratio of 0.05.
We save and evaluate the model for every epoch, and the best results are recorded.
For the video training, we use a more expressive LoRA configuration with a rank of 256 and an alpha of 512, training for a single epoch with a learning rate of 5e-6. 
Across all experiments, we use the AdamW optimizer, and all training is conducted on bf16.

\subsection{Results and Analysis}

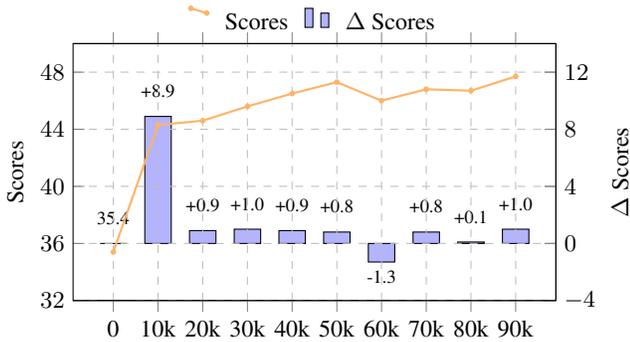
\begin{figure}[t!]
    \centering\small
    \begin{tikzpicture}

        \pgfplotscreateplotcyclelist{mycolors}{
            {fill=blue!30}
        }

        \begin{axis}[
            axis y line*=right,
            ymin=-4,
            ymax=14,
            ylabel={$\Delta$ Scores},
            ytick={-4,0,4,8,12},
            grid=none, 
            enlarge x limits=0.1,
            symbolic x coords={0,10k,20k,30k,40k,50k,60k,70k,80k,90k},
            xtick=\empty, 
            width=8cm,
            height=5cm,
            cycle list name=mycolors,
            nodes near coords, % Simply enable nodes near coords
            point meta=explicit symbolic,
            every node near coord/.append style={
                font=\scriptsize,
                yshift=4pt,
                anchor=south,
                color=black %
            }
        ]
        \addplot+[ybar, bar shift=0pt] coordinates {(0,0) [35.4]}; 
        \addplot+[ybar, bar shift=0pt] coordinates {(10k,8.9) [+8.9]}; 
        \addplot+[ybar, bar shift=0pt] coordinates {(20k,0.9) [+0.9]}; 
        \addplot+[ybar, bar shift=0pt] coordinates {(30k,1.0) [+1.0]}; 
        \addplot+[ybar, bar shift=0pt] coordinates {(40k,0.9) [+0.9]}; 
        \addplot+[ybar, bar shift=0pt] coordinates {(50k,0.8) [+0.8]}; 
        % \addplot+[ybar, bar shift=0pt] coordinates {(60k,-1.3) [-1.3]}; 
        \addplot+[ybar, bar shift=0pt,
            every node near coord/.append style={yshift=-4pt, anchor=north} 
        ] coordinates {(60k,-1.3) [-1.3]};
        \addplot+[ybar, bar shift=0pt] coordinates {(70k,0.8) [+0.8]}; 
        \addplot+[ybar, bar shift=0pt] coordinates {(80k,0.1) [+0.1]}; 
        \addplot+[ybar, bar shift=0pt] coordinates {(90k,1.0) [+1.0]}; 

        \end{axis}

        \begin{axis}[
            ybar, 
            bar width=12pt, 
            ymin=32,
            ymax=50,
            ylabel={Scores},
            ytick={32,36,40,44,48},
            grid=major,
            grid style={dashed, gray!50},
            enlarge x limits=0.1, 
            legend style={
                at={(0.5,1)},
                anchor=south,
                column sep=1ex,
                draw=none,
                fill=none,
                legend columns=-1
            },
            symbolic x coords={0,10k,20k,30k,40k,50k,60k,70k,80k,90k},
            xtick={0,10k,20k,30k,40k,50k,60k,70k,80k,90k},
            xticklabel style={
                rotate=0, 
                anchor=north,
                font=\small
            },
            width=8cm,
            height=5cm,
            cycle list name=mycolors,
            nodes near coords, % Simply enable nodes near coords
            point meta=explicit symbolic,
            every node near coord/.append style={
                font=\scriptsize,
                yshift=4pt,
                anchor=south,
                color=red!50 %
            }
        ]

        \addplot[
            sharp plot,
            color=orange!60,
            thick,
            mark=asterisk,
            mark options={scale=0.5, fill=blue, line width=1pt},
            nodes near coords={},
            point meta=x
        ] coordinates {
            (0,35.4)
            (10k,44.3)
            (20k,44.6)
            (30k,45.6)
            (40k,46.5)
            (50k,47.3)
            (60k,46.0)
            (70k,46.8)
            (80k,46.7)
            (90k,47.7)
        };
        \addlegendentry{Scores}
        \addplot+[ybar, bar shift=0pt] coordinates {(0,0) [35.4]}; \addlegendentry{$\Delta$ Scores}
        \label{plots:line} 
        \end{axis}

    \end{tikzpicture}
    \caption{Performance trends under increasing training data scales for video-based scenarios. Here, $\Delta$ Scores denotes the change in performance compared to the preceding training data scale.}
    \label{fig:data_train_video}
\end{figure}

Figure~\ref{fig:data_train_multiview_wholefig} and Figure~\ref{fig:data_train_video} present the performance trends under increasing training data scales for multi-view and video-based scenarios, respectively. We observe three key findings:

\textbf{Limited Data Gains.} 
Model performance improves with increased training data, consistent with established scaling laws. However, the performance gains saturate quickly, and the overall improvement ceiling remains relatively low. 
This suggests that while more data is beneficial, its marginal utility diminishes rapidly for these complex spatial tasks, converging faster and at a lower cap than typically seen in general-domain scaling.

\textbf{Task Type Matters.}
The performance ceiling is not uniform across all tasks; it is intrinsically linked to task type.
As shown in Figure~\ref{fig:data_train_multiview_wholefig}, occlusion restoration achieves higher accuracy with less data compared to azimuth transfer.
While occlusion restoration can be effectively addressed through semantic matching to understand the relationships between multiple views, the azimuth transfer task necessitates both spatial imagination and a deeper understanding of spatial relationships. 
The comparison of distances requires an intrinsic ability for spatial perception.
It indicated that MLLMs are limited in spatial intuition and imaginative capabilities.
A detailed breakdown of performance across each task on the VSI-bench is provided in the Appendix.

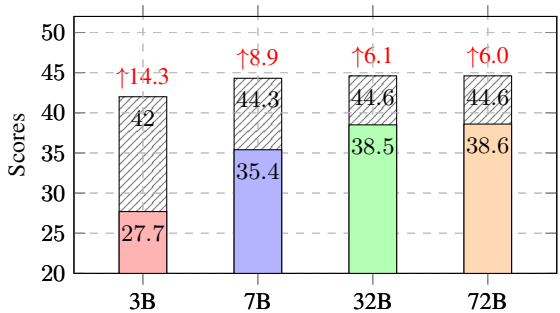
\begin{figure}[t!]

\centering\small
\begin{tikzpicture}

      \pgfplotscreateplotcyclelist{mycolors}{
            {fill=red!30},
            {fill=blue!30},
            {fill=green!30},
            {fill=orange!30}}
        \begin{axis}[
            ybar, 
            bar width=18pt, 
            ymin=20,
            ymax=52,
            ylabel={Scores},
            ytick={20,25,30,35,40,45,50},
            grid=major,
            grid style={dashed, gray!50},
            enlarge x limits=0.2, 
            legend style={
                at={(0.5,1)},
                anchor=south,
                column sep=1ex,
                draw=none,
                fill=none,
                legend columns=-1
            },
            symbolic x coords={3B,7B,32B,72B},
            xtick={3B,7B,32B,72B},
            xticklabel style={
                rotate=0, 
                anchor=north,
                font=\small
            },
            width=8cm,
            height=5cm,
            cycle list name=mycolors
        ]
        
        \addplot[ybar, 
                    bar shift=0pt, 
                    nodes near coords,  
                    nodes near coords style={
                    font=\small,
                    color=black,
                    anchor=north, 
                    inner sep=5pt},
                    fill=gray!30, 
                    pattern=north east lines, 
                    pattern color=gray] coordinates {(3B,42)}; 
        % \addplot[ybar, bar shift=0pt,fill=gray!30, pattern=north east lines, pattern color=gray,draw=black]coordinates {(3B,40)};
        \addplot[ybar, 
                    bar shift=0pt, 
                    nodes near coords,  
                    nodes near coords style={
                    font=\small,
                    color=black,
                    anchor=north, 
                    inner sep=5pt},
                    fill=gray!30, 
                    pattern=north east lines, 
                    pattern color=gray] coordinates {(7B,44.3)};
        \addplot[ybar, 
                    bar shift=0pt, 
                    nodes near coords,  
                    nodes near coords style={
                    font=\small,
                    color=black,
                    anchor=north, 
                    inner sep=5pt},
                    fill=gray!30, 
                    pattern=north east lines, 
                    pattern color=gray]coordinates {(32B,44.6)};
        \addplot[ybar, 
                    bar shift=0pt, 
                    nodes near coords,  
                    nodes near coords style={
                    font=\small,
                    color=black,
                    anchor=north, 
                    inner sep=5pt},
                    fill=gray!30, 
                    pattern=north east lines, 
                    pattern color=gray]coordinates {(72B,44.6)};
    \node[red, font=\small, above] at (axis cs:3B, 42) {$\uparrow$14.3};
    \node[red, font=\small, above] at (axis cs:7B, 44.3) {$\uparrow$8.9}; 
    \node[red, font=\small, above] at (axis cs:32B, 44.6) {$\uparrow$6.1};
    \node[red, font=\small, above] at (axis cs:72B, 44.6) {$\uparrow$6.0};

    \end{axis}

        \begin{axis}[
            ybar, 
            bar width=18pt, 
            ymin=20,
            ymax=52,
            ytick={20,25,30,35,40,45,50},
            grid=major,
            grid style={dashed, gray!50},
            enlarge x limits=0.2, 
            legend style={
                at={(0.5,1)},
                anchor=south,
                column sep=1ex,
                draw=none,
                fill=none,
                legend columns=-1
            },
            symbolic x coords={3B,7B,32B,72B},
            xtick={3B,7B,32B,72B},
            xticklabel style={
                rotate=0, 
                anchor=north,
                font=\small
            },
            width=8cm,
            height=5cm,
        ]

        \addplot[ybar, 
                    bar shift=0pt, 
                    fill = red!30,
                    nodes near coords,  
                    nodes near coords style={
                    font=\small,
                    color=black,
                    anchor=north,
                    inner sep=5pt
                    }] coordinates {(3B,27.7)};
        \addplot[ybar,
                    fill = blue!30,
                    bar shift=0pt, 
                    nodes near coords,  
                    nodes near coords style={
                    font=\small,
                    color=black,
                    anchor=north, 
                    inner sep=5pt
                    }] coordinates {(7B,35.4)};
        \addplot[ybar, 
                    fill = green!30,
                    bar shift=0pt, 
                    nodes near coords,  
                    nodes near coords style={
                    font=\small,
                    color=black,
                    anchor=north, 
                    inner sep=5pt
                    }]  coordinates {(32B,38.5)};
        \addplot[ybar, 
                    fill = orange!30,
                    bar shift=0pt, 
                    nodes near coords,  
                    nodes near coords style={
                    font=\small,
                    color=black,
                    anchor=north,
                    inner sep=5pt
                    }]  coordinates {(72B,38.6)};
   
        \end{axis}
\end{tikzpicture}\vspace{-3pt}
\caption{Scores for different models (Base Score + Fine-tune Increment)}
\label{fig:model_size_train}\vspace{-6mm}
\end{figure}

\textbf{Model size matters.}
To isolate the impact of model scale, we fine-tuned a family of Qwen2.5-VL (3B, 7B, 32B, and 72B) on a consistent subset of 10,000 samples from the SpaceR-151k dataset. As shown in Figure~\ref{fig:model_size_train}, our results show a clear trend: while larger models start from a higher baseline, the performance gains from fine-tuning decrease as model size increases. This pattern of diminishing returns suggests that for specialized tasks like spatial reasoning, simply increasing model parameters is not a panacea. 

In summary, our data-centric analysis reveals that simply scaling up the training data volume provides diminishing returns for enhancing MLLMs' spatial understanding capabilities. 
And the upper bound is relatively low, especially for tasks that require spatial imagination.

\section{Architecture-Centric Analysis}
\label{sec:architecture_analysis}

%==============================================================
\begin{table*}[t]
\centering\small
% \resizebox{\linewidth}{!}{%
\setlength\tabcolsep{7pt}
\caption{Ablation analysis of positional encodings (\#1 for Qwen-2.5-VL-7B, \#2 for LLaVA-OV-7B, and \#3 for Mono-InternVL-2B); (*) indicates both text and vision tokens are modified, while others modify only text tokens. Each ablation type follows the format: \texttt{strategy-architecture-dimension}. Note that multi-view experiments were omitted for Mono-InternVL-2B, as its architecture is designed for single-image inputs.}
\begin{tabular}{llcccccccc}
\toprule
\multirow{4}{*}{\textbf{}} & \multirow{4}{*}{\textbf{Ablation Type}} 
& \multicolumn{3}{c}{\textbf{Single-View}} & \multicolumn{3}{c}{\textbf{Multi-View}} \\
\cmidrule(lr){3-5} \cmidrule(lr){6-8}
& & \makecell{\textbf{What'sUp}} 
& \makecell{\textbf{COCO\_QA}} 
& \makecell{\textbf{VG} \\ \textbf{Attribution}}
& \makecell{\textbf{Occlusion} \\ \textbf{Restoration}} 
& \makecell{\textbf{Distance} \\ \textbf{Comparison}} 
& \makecell{\textbf{Azimuth} \\ \textbf{Transfer}} \\
\midrule

\multirow{10}{*}{\#1} %
& Vanilla         & 86.65 & 87.32 & 90.79 & 34.52 & 43.41 & 27.25 \\
& Shuffle-LLM-txy(*) & 2.92 (-83.73) & 0.96 (-86.36) & 1.63 (-89.16) & 0.41 (-34.11) & 0.85 (-42.56) & 0.61 (-26.64) \\
& Shuffle-LLM-xy(*)  & 80.79 (-5.86) & 76.77 (-10.55) & 90.61 (-0.18) & 26.64 (-7.88) & 38.84 (-4.57) & 30.54 (+3.29) \\
& Shuffle-LLM-txy & 81.16 (-5.49) & 81.40 (-5.92) & 87.92 (-2.87) & 37.19 (+2.67) & 39.01 (-4.40) & 33.78 (+6.53) \\
& Shuffle-LLM-xy & 89.48 (+2.83) & 85.13 (-2.19) & 90.62 (-0.17) & 33.83 (-0.69) & 43.14 (-0.27) & 27.86 (+0.61) \\
& Mask-LLM-txy  & 90.52 (+3.87) & 86.45 (-0.87) & 89.45 (-1.34) & 24.96 (-9.56) & 42.66 (-0.75) & 29.54 (+2.29) \\
& Mask-LLM-xy   & 89.55 (+2.90) & 86.39 (-0.93) & 90.59 (-0.20) & 34.03 (-0.49) & 42.70 (-0.71) & 28.36 (+1.11) \\
& Shuffle-VE-hw  & 8.99 (-77.66) & 34.93 (-52.39) & 85.85 (-4.94) & 22.47 (-12.05) & 36.89 (-6.52) & 30.71 (+3.46) \\
& Shuffle-VE-h   & 49.28 (-37.37) & 61.44 (-25.88) & 87.13 (-3.66) & 30.35 (-4.17) & 36.86 (-6.55) & 33.05 (+5.80) \\
& Mask-VE-h      & 26.03 (-60.62) & 66.21 (-21.11) & 84.63 (-6.16) & 25.36 (-9.16) & 33.48 (-9.93) & 33.28 (+6.03) \\

\midrule
\multirow{5}{*}{\#2} 
& Vanilla               & 98.66 & 86.61 & 94.33 & 22.38 & 43.75 & 24.67 \\
& Shuffle-LLM-txy       & 97.68 (-0.98) & 72.52 (-14.09) &  94.27 (-0.06) & 18.82 (-3.56) & 41.94 (-1.81) & 22.50 (-2.17) \\
& Const-LLM-First       & 97.81 (-0.85) & 69.47 (-17.14) &  96.42 (+2.09) & 19.80 (-2.58) & 40.10 (-3.65) & 23.05 (-1.62) \\
& Const-LLM-Last        & 0.00 (-98.66) & 41.86 (-44.75) & 85.54 (-8.79) & 22.35 (-0.03) & 35.52 (-8.23) & 21.88 (-2.79) \\
& Shuffle-VE-hw         & 12.31 (-86.35) & 23.91 (-62.70) & 95.84 (+1.51) & 16.36 (-6.02) & 35.11 (-8.64) & 22.44 (-2.23) \\

\midrule
\multirow{5}{*}{\#3} 
& Vanilla               & 37.45 & 31.26 & 80.09 & - & - & - \\
& Shuffle-LLM-txy       & 9.53 (-27.92) & 18.12 (-13.14) &  75.25 (-4.84) & - & - & - \\
& Const-LLM-First       & 23.14 (-14.31) & 29.12 (-2.14) &  80.77 (+0.68) & - & - & - \\
& Const-LLM-Last        & 10.73 (-26.72) & 11.59 (-19.67) &  26.04 (-54.05) & - & - & - \\
& Shuffle-VE-hw         & 22.47 (-14.98) & 16.32 (-14.94) &  72.79 (-7.30) & - & - & - \\

\bottomrule
\end{tabular}
\label{tab:ablation_PE}
\end{table*}

%==============================================================

In Transformer-based architectures, \textbf{P}ositional \textbf{E}ncoding (PE) is the fundamental mechanism for processing spatial information. The representation of ``where'' an object is, both in the 2D image plane and in the 1D sequence of tokens, is entirely dependent on PE. To investigate the precise impact of PE on spatial understanding, we conduct a series of ablation studies on its role, spanning from single-view to multi-view scenarios and from the vision encoders to the language modules.

\subsection{Preliminaries}

Positional encodings are crucial for MLLMs to interpret spatial and sequential data. The cascaded MLLM Qwen2.5-VL~\cite{bai2025qwen25vltechnicalreport} utilizes a comprehensive RoPE-based strategy, with 2D-RoPE~\cite{heo2024rotary} in its \textbf{V}ision \textbf{E}ncoder (VE) and M-RoPE~\cite{bai2025qwen25vltechnicalreport} in its LLM. In contrast, other prominent models, including the cascaded MLLM LLaVA-OneVision~\cite{li2024llavaov} and the native MLLM Mono-InternVL~\cite{luo2025mono}, adopt a more conventional setup: they pair a VE using learnable PEs with an LLM that employs standard 1D-RoPE~\cite{su2024roformer}. This diversity in PE configurations motivates our subsequent analysis of their dimensional properties.

\subsection{Experiment Setup}
\textbf{Model and Dataset.} We evaluate the cascaded MLLM Qwen2.5-VL, LLaVA-OneVision, and native MLLM Mono-InternVL on a range of benchmarks.  
For single-view understanding, we use What'sUp~\cite{kamath-etal-2023-whats} and COCO-QA~\cite{lin2014microsoft} to evaluate spatial relationships tasks, supplemented by VG-Attribute~\cite{krishna2017visual} for non-spatial tasks.
For multi-view spatial understanding, we use our proposed MulSeT benchmark.For inference, we use a batch size of 256 and a fixed seed of 1 for reproducibility.

\textbf{Ablation Strategy.} 
Our ablation study includes three distinct strategies to investigate the role of positional encodings (PEs). 
The \textbf{Mask} strategy nullifies a specific PE dimension by setting all to zero. 
The \textbf{Shuffle} strategy disrupts positional information by randomly permuting the positional indices within a targeted dimension, which preserves the original value distribution. For \textbf{Constant Value} strategy, we neutralize a dimension by assigning all its positions the value from a single reference token (e.g., the first or last token in the sequence). Furthermore, we control the scope of these ablations by specifying whether they apply to the text or image modality and by selecting a precise target PE dimension.

\subsection{Result and Analysis}

\textbf{The positional encoding from the Vision Encoder plays an important role in spatial understanding.}
As shown in Fig~\ref{tab:ablation_PE}, positional information is a critical determinant of a model's spatial capabilities. 
The MLLMs exhibit a far greater dependency on the positional encoding from the Vision Encoder (VE) than from the LLM. 
Disabling the VE's RoPE leads to a catastrophic decline in performance, establishing the VE's explicit 2D positional signals as the primary foundation for spatial understanding in these models.

\textbf{Positional encodings in different architectural models have distinct characteristics in spatial understanding.} 
Cascaded models such as LLaVA-OneVision demonstrate a stronger reliance on the VE's positional information compared to natively integrated models like Mono-InternVL. Interestingly, this trend reverses for the non-spatial VG attribute recognition task, where the performance of the cascaded model remains almost entirely unaffected by RoPE modifications, suggesting a more intricate fusion of visual and textual features in end-to-end trained architectures.

Further investigation into the LLM's positional signals shows that setting all visual token positions to that of the last token (Const-LLM-Last) incurs a significantly greater performance penalty than using the first token's position. We attribute this to the common training strategy of appending a global image thumbnail feature after the local image patches, making the positional context of this final token uniquely important. Notably, as detailed in the Appendix, our directional evaluations on single-view tasks confirm this mechanism, showing that the height ($h$) and width ($w$) components of the VE's RoPE directly correlate with the model's performance on vertical and horizontal discrimination tasks.

\section{Solution Exploration}

\label{sec:solution_exploration}
 
Recent studies have explored enhancing the spatial understanding, including reasoning injection~\cite{jin2025wellthinkingenhancingllm} and architectural improvements~\cite{wu2025spatialmllmboostingmllmcapabilities}. Reasoning injection refers to eliciting spatial reasoning capabilities through carefully designed prompting strategies without altering the model's structure. Architectural improvements seek to embed stronger spatial inductive biases directly into the model's architecture. In the subsequent subsections, we provide a comprehensive analysis of both methodologies.

\subsection{Reasoning Injection}
\label{sec:reasoning_injection}
%====================================================================
\begin{table*}[t!]
\centering
\small
\caption{Performance comparison of different inference methods on MulSeT. The best and second-best results are marked in \textbf{bold} and \underline{underlined}, respectively.}
\begin{tabular}{llccc}
\toprule
\textbf{Model} & \textbf{Inference Method} & \textbf{Occlusion Restoration} & \textbf{Distance Comparison} & \textbf{Azimuth Transfer} \\
\midrule
Qwen2.5-VL & Vanilla & 34.52 & \textbf{43.41} & 27.25 \\
 & Implicit Stepwise & \underline{35.04}(+0.52) & 42.42(-0.99) & \underline{28.25}(+1.00) \\
 & Implicit Multi-view & \textbf{36.37}(+1.85) & \underline{42.49}(-0.92) & \textbf{29.15}(+1.90) \\
 & Explicit Stepwise CoT & 33.01(-1.51) & 34.41(-9.00) & 23.00(-4.25) \\
 & Explicit Multi-view CoT & \underline{35.04}(+0.52) & 34.16(-9.25) & 24.29(-2.96) \\
\bottomrule
\end{tabular}
\label{tab:cot_comparison_percent}
\end{table*}

%====================================================================

%====================================================================
\begin{figure}[t]
\centering
\includegraphics[width=1\linewidth]{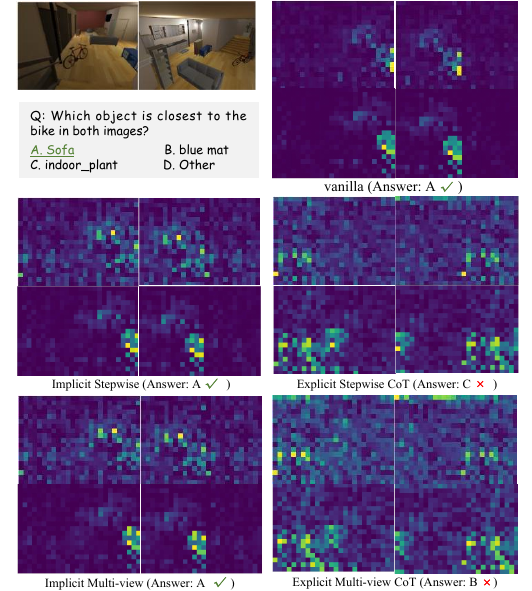}
\caption{Case study of attention visualization. Each column corresponds to a pair of input images (left and right), and each row shows the attention maps from the 21st and 27th layers of the model, respectively.}
\vspace{-6mm}
\label{fig:attention}
\end{figure}
%====================================================================

Reasoning injection can be broadly divided into two categories based on the output format: explicit and implicit reasoning. Explicit reasoning requires the model to generate its intermediate thought process as part of the output, with Chain-of-Thought (CoT)~\cite{wei2022chain} being the most prominent example. Numerous studies have applied this to spatial reasoning, designing structured CoT prompts to decompose complex problems~\cite{0001W0ZZLH24, li2025imaginereasoningspacemultimodal,liao2025improvedvisualspatialreasoningr1zerolike,yang2025thinkingspacemultimodallarge, feng2025videor1reinforcingvideoreasoning, wu2025groundedchainofthoughtmultimodallarge, liu2025spatialcotadvancingspatialreasoning}. In contrast, implicit reasoning guides the model's internal cognitive process through carefully formulated prompts~\cite{jin2025wellthinkingenhancingllm,lin2025evo}. 
% Given the diverse strategies and mixed results, particularly for explicit methods, a systematic analysis is warranted. 
In the subsequent sections, we will evaluate representative strategies from both categories to gain a better exploration of their efficacy.

\textbf{Experiment Setup.}
Inspired by the human cognitive process of continually comparing images to understand spatial changes, we investigate the impact of different reasoning injection strategies. We designed four prompt variants structured along two axes: a standard stepwise instruction versus our proposed multi-view consistency instruction, and an implicit versus an explicit reasoning format. The core difference between the implicit and explicit versions lies in a single concluding sentence that either directs the model to reason privately or to output its thought process. The specific templates are detailed in Table~\ref{tab:prompt_templates}.

Our experiments are conducted using the Qwen2.5-VL-7B model on MulSeT dataset. All evaluations are performed with a batch size of 256 and a fixed seed of 1 to ensure reproducibility.

\begin{table}[t!]
\centering
\small
\caption{Prompt templates for reasoning injection.}
\begin{tabular}{p{0.9\linewidth}}
\hline
\textbf{Implicit Stepwise Prompt} \\ \hline
\textit{Let's think step by step before answering the question! And you should perform your step-by-step reasoning privately and not reveal it to the user.} \\ \hline
% \\
\hline
\textbf{Implicit Multi-View Consistency Prompt} \\ \hline
\textit{Before answering, compare the two images to identify the same objects across different viewpoints. Analyze how each object's appearance, position, and visibility change due to the viewpoint shift. Use this comparison to resolve spatial ambiguities. Think step by step. Ensure your reasoning is consistent and grounded in visual evidence from all views. Only after this reasoning, choose the correct answer. And you should perform your step-by-step reasoning privately and not reveal it to the user.} \\ \hline 
% \\
\hline
\textbf{Explicit Reasoning Prompt } \\ \hline
Replace the final sentence of the above Prompt with the following: \textit{Please briefly show your reasoning process (within 500 words if possible), and conclude your answer in the format: 'The answer is A/B/C/D'.} \\ \hline
\end{tabular}
\label{tab:prompt_templates}
\end{table}

\textbf{Performance Analysis.}
The experimental results, as detailed in Table~\ref{tab:cot_comparison_percent}, reveal three key findings. First, the efficacy of reasoning injection is highly correlated with task complexity. For tasks demanding significant logical inference, such as occlusion restoration and azimuth transfer, reasoning injection yields consistent performance gains. Conversely, for tasks like distance comparison that rely more on the model's intrinsic perceptual capabilities, the injection process can be detrimental to performance. Second, for all three multi-view spatial understanding tasks, our reasoning injection tailored for multi-view contexts consistently outperforms the traditional stepwise approach. This underscores the importance of task-specific prompting for achieving optimal results. Finally, we observe a substantial performance degradation when employing explicit reasoning compared to its implicit counterpart. This pronounced disparity motivates a deeper investigation into the model's internal mechanisms. Therefore, in the following section, we conduct an attention visualization analysis to better understand the underlying causes of these performance differences.

\textbf{Attention Visualization Analysis.}
As illustrated in Figure~\ref{fig:attention}, the attention maps of the Explicit model are considerably more diffused than those of the Implicit model. This indicates that the explicit reasoning process alters the model's attentional focus. In the specific case shown, identifying the object closest to the bike requires discriminating between the nearby sofa and TV stand. Both the vanilla and Implicit models correctly focus on these objects. In contrast, the Explicit model concentrates its attention primarily on the bike itself, overlooking the critical surrounding context. We follow the methodology proposed by~\citet{zhang2025mllms} to visualize the attention maps.
 Comprehensive layer-by-layer visualizations are provided in the Appendix.

\subsection{Architectural Enhancements}
\label{sec:architectural_enhancements}

Architectural enhancements aim to embed stronger spatial inductive biases directly into the model, either by \textbf{introducing new components}~\cite{wu2025spatialmllmboostingmllmcapabilities, zheng2025video3dllmlearningpositionaware} or by \textbf{modifying existing ones}~\cite{wei2025videoropemakesgoodvideo}.
For instance, some works introduce new modules like a dual-encoder for disentangled processing~\cite{wu2025spatialmllmboostingmllmcapabilities} or add explicit 3D positional information to features~\cite{zheng2025video3dllmlearningpositionaware}.
Others modify existing components, such as adapting rotary positional encodings with a 3D structure to better capture spatio-temporal relationships in general video tasks~\cite{wei2025videoropemakesgoodvideo}.

Motivated by our analysis in Sec~\ref{sec:architecture_analysis}, we posit that positional encodings require more specialized refinement for spatial understanding. A more promising architectural direction is to better preserve and integrate the VE's rich spatial information into the language model's reasoning process.

\section{Related Work}

\textbf{Architectures of MLLMs.}
Existing MLLMs can be broadly categorized into two architectural paradigms: cascaded and native. 
Cascaded MLLMs employ a modular pipeline, connecting a pre-trained visual encoder (e.g., ViT~\cite{dosovitskiy2021imageworth16x16words}, CLIP~\cite{radford2021learningtransferablevisualmodels}) to an LLM via a lightweight projection layer or adapter. 
This design, exemplified by influential models such as LLaVA~\cite{liu2023visual}, InstructBLIP~\cite{dai2023instructblipgeneralpurposevisionlanguagemodels}, Flamingo~\cite{alayrac2022flamingo}, and GPT-4~\cite{achiam2023gpt}, effectively leverages powerful, independently components. 
In contrast, native MLLMs, also referred to as monolithic or encoder-free models, pursue a more unified architecture. These models deeply integrate visual perception capabilities directly into the LLM backbone~\cite{luo2025mono,diao2025EVEv2}.

\textbf{Analysis of Spatial Understanding Failures in MLLMs.}
Recent efforts have begun to probe the limitations of MLLMs in spatial understanding, often through component-level or phenomenon-driven analyses. For example, some studies attribute failures to the biases of pre-trained vision encoders such as CLIP, which may lead to systematic spatial errors~\cite{tong2024eyes}. Others investigate attention patterns, finding that improper focus allocation correlates with poor spatial reasoning, particularly in single-view scenarios~\cite{chen2025spatialreasoninghardvlms}. \citet{yang2025thinkingspacemultimodallarge} propose that building internal cognitive maps enhances distance reasoning capabilities, while~\citet{chen2024multiobjecthallucinationvisionlanguagemodels} conduct failure case analyses highlighting issues like object hallucination and relational confusion. 
Despite these advances, our work aims to bridge this gap through a comprehensive analysis of spatial understanding.

\section{Conclusion}
In this work, we presented a systematic analysis of the spatial understanding capabilities of MLLMs from both data and architectural perspectives across three representative scenarios: single-view, multi-view, and videos. And as a core component, we introduced a novel benchmark, MulSeT, specifically for multi-view spatial tasks. Through a series of targeted experiments, our findings reveal two key conclusions: first, from a data standpoint, merely scaling up training data is insufficient to significantly improve spatial understanding, as MLLMs exhibit a low performance upper bound on these tasks. Second, from an architectural standpoint, we discovered that the positional encoding within the visual encoder is substantially more critical for spatial understanding than that in the language model. These insights underscore the fundamental limitations of current MLLMs and suggest that future progress in spatial reasoning will necessitate targeted reasoning injection and architectural innovations rather than relying on data scaling alone.
 
\bibliography{main}

\begin{thebibliography}{47}
\providecommand{\natexlab}[1]{#1}

\bibitem[{Achiam et~al.(2023)Achiam, Adler, Agarwal, Ahmad, Akkaya, Aleman, Almeida, Altenschmidt, Altman, Anadkat et~al.}]{achiam2023gpt}
Achiam, J.; Adler, S.; Agarwal, S.; Ahmad, L.; Akkaya, I.; Aleman, F.~L.; Almeida, D.; Altenschmidt, J.; Altman, S.; Anadkat, S.; et~al. 2023.
\newblock Gpt-4 technical report.
\newblock \emph{arXiv preprint arXiv:2303.08774}.

\bibitem[{Alayrac et~al.(2022)Alayrac, Donahue, Luc, Miech, Barr, Hasson, Lenc, Mensch, Millican, Reynolds et~al.}]{alayrac2022flamingo}
Alayrac, J.-B.; Donahue, J.; Luc, P.; Miech, A.; Barr, I.; Hasson, Y.; Lenc, K.; Mensch, A.; Millican, K.; Reynolds, M.; et~al. 2022.
\newblock Flamingo: a visual language model for few-shot learning.
\newblock \emph{Neural Information Processing Systems}.

\bibitem[{Bai et~al.(2025)Bai, Chen, Liu, Wang, Ge, Song, Dang, Wang, Wang, Tang, Zhong, Zhu, Yang, Li, Wan, Wang, Ding, Fu, Xu, Ye, Zhang, Xie, Cheng, Zhang, Yang, Xu, and Lin}]{bai2025qwen25vltechnicalreport}
Bai, S.; Chen, K.; Liu, X.; Wang, J.; Ge, W.; Song, S.; Dang, K.; Wang, P.; Wang, S.; Tang, J.; Zhong, H.; Zhu, Y.; Yang, M.; Li, Z.; Wan, J.; Wang, P.; Ding, W.; Fu, Z.; Xu, Y.; Ye, J.; Zhang, X.; Xie, T.; Cheng, Z.; Zhang, H.; Yang, Z.; Xu, H.; and Lin, J. 2025.
\newblock Qwen2.5-VL Technical Report.
\newblock arXiv:2502.13923.

\bibitem[{Chen et~al.(2025)Chen, Zhu, Zhou, Zhang, Gao, Niebles, Geva, He, Wu, and Li}]{chen2025spatialreasoninghardvlms}
Chen, S.; Zhu, T.; Zhou, R.; Zhang, J.; Gao, S.; Niebles, J.~C.; Geva, M.; He, J.; Wu, J.; and Li, M. 2025.
\newblock Why Is Spatial Reasoning Hard for VLMs? An Attention Mechanism Perspective on Focus Areas.
\newblock In \emph{Proceedings of the International Conference on Machine Learning}.

\bibitem[{Chen et~al.(2024)Chen, Ma, Zhang, Xu, Qian, Yang, Fouhey, and Chai}]{chen2024multiobjecthallucinationvisionlanguagemodels}
Chen, X.; Ma, Z.; Zhang, X.; Xu, S.; Qian, S.; Yang, J.; Fouhey, D.~F.; and Chai, J. 2024.
\newblock Multi-Object Hallucination in Vision-Language Models.
\newblock arXiv:2407.06192.

\bibitem[{Cheng et~al.(2025)Cheng, Tu, Li, Dai, Hu, Hu, Li, Shi, Yu, Chen, Shi, and Sun}]{cheng2025embodiedevalevaluatemultimodalllms}
Cheng, Z.; Tu, Y.; Li, R.; Dai, S.; Hu, J.; Hu, S.; Li, J.; Shi, Y.; Yu, T.; Chen, W.; Shi, L.; and Sun, M. 2025.
\newblock EmbodiedEval: Evaluate Multimodal LLMs as Embodied Agents.
\newblock In \emph{The Annual Meeting of the Association for Computational Linguistics}.

\bibitem[{Colan, Davila, and Hasegawa(2025)}]{colan2025assessingvaluevisualinput}
Colan, J.; Davila, A.; and Hasegawa, Y. 2025.
\newblock Assessing the Value of Visual Input: A Benchmark of Multimodal Large Language Models for Robotic Path Planning.
\newblock arXiv:2507.12391.

\bibitem[{Dai et~al.(2023)Dai, Li, Li, Tiong, Zhao, Wang, Li, Fung, and Hoi}]{dai2023instructblipgeneralpurposevisionlanguagemodels}
Dai, W.; Li, J.; Li, D.; Tiong, A. M.~H.; Zhao, J.; Wang, W.; Li, B.; Fung, P.; and Hoi, S. 2023.
\newblock InstructBLIP: Towards General-purpose Vision-Language Models with Instruction Tuning.
\newblock arXiv:2305.06500.

\bibitem[{Diao et~al.(2025)Diao, Li, Cui, Wang, Deng, Pan, Wang, Lu, and Wang}]{diao2025EVEv2}
Diao, H.; Li, X.; Cui, Y.; Wang, Y.; Deng, H.; Pan, T.; Wang, W.; Lu, H.; and Wang, X. 2025.
\newblock EVEv2: Improved Baselines for Encoder-Free Vision-Language Models.
\newblock \emph{arXiv preprint arXiv:2502.06788}.

\bibitem[{Dosovitskiy et~al.(2021)Dosovitskiy, Beyer, Kolesnikov, Weissenborn, Zhai, Unterthiner, Dehghani, Minderer, Heigold, Gelly, Uszkoreit, and Houlsby}]{dosovitskiy2021imageworth16x16words}
Dosovitskiy, A.; Beyer, L.; Kolesnikov, A.; Weissenborn, D.; Zhai, X.; Unterthiner, T.; Dehghani, M.; Minderer, M.; Heigold, G.; Gelly, S.; Uszkoreit, J.; and Houlsby, N. 2021.
\newblock An Image is Worth 16x16 Words: Transformers for Image Recognition at Scale.
\newblock arXiv:2010.11929.

\bibitem[{Du et~al.(2024)Du, Wu, Li, Huang, and Wei}]{du-etal-2024-embspatial}
Du, M.; Wu, B.; Li, Z.; Huang, X.; and Wei, Z. 2024.
\newblock {E}mb{S}patial-Bench: Benchmarking Spatial Understanding for Embodied Tasks with Large Vision-Language Models.
\newblock In Ku, L.-W.; Martins, A.; and Srikumar, V., eds., \emph{Proceedings of the 62nd Annual Meeting of the Association for Computational Linguistics (Volume 2: Short Papers)}, 346--355. Bangkok, Thailand: Association for Computational Linguistics.

\bibitem[{Fei et~al.(2024)Fei, Wu, Ji, Zhang, Zhang, Lee, and Hsu}]{0001W0ZZLH24}
Fei, H.; Wu, S.; Ji, W.; Zhang, H.; Zhang, M.; Lee, M.-L.; and Hsu, W. 2024.
\newblock Video-of-Thought: Step-by-Step Video Reasoning from Perception to Cognition.
\newblock In \emph{Proceedings of the International Conference on Machine Learning}.

\bibitem[{Feng et~al.(2025)Feng, Gong, Li, Guo, Wang, Peng, Wu, Zhang, Wang, and Yue}]{feng2025videor1reinforcingvideoreasoning}
Feng, K.; Gong, K.; Li, B.; Guo, Z.; Wang, Y.; Peng, T.; Wu, J.; Zhang, X.; Wang, B.; and Yue, X. 2025.
\newblock Video-R1: Reinforcing Video Reasoning in MLLMs.
\newblock arXiv:2503.21776.

\bibitem[{Heo et~al.(2024)Heo, Park, Han, and Yun}]{heo2024rotary}
Heo, B.; Park, S.; Han, D.; and Yun, S. 2024.
\newblock Rotary position embedding for vision transformer.
\newblock In \emph{European Conference on Computer Vision}, 289--305. Springer.

\bibitem[{Hu et~al.(2022)Hu, Shen, Wallis, Allen-Zhu, Li, Wang, Wang, Chen et~al.}]{hu2022lora}
Hu, E.~J.; Shen, Y.; Wallis, P.; Allen-Zhu, Z.; Li, Y.; Wang, S.; Wang, L.; Chen, W.; et~al. 2022.
\newblock Lora: Low-rank adaptation of large language models.
\newblock \emph{International Conference on Learning Representations}, 1(2): 3.

\bibitem[{Hurst et~al.(2024)Hurst, Lerer, Goucher, Perelman, Ramesh, Clark, Ostrow, Welihinda, Hayes, Radford et~al.}]{hurst2024gpto}
Hurst, A.; Lerer, A.; Goucher, A.~P.; Perelman, A.; Ramesh, A.; Clark, A.; Ostrow, A.; Welihinda, A.; Hayes, A.; Radford, A.; et~al. 2024.
\newblock GPT-4o System Card.
\newblock \emph{arXiv preprint arXiv:2410.21276}.

\bibitem[{Jia et~al.(2025)Jia, Qi, Zhang, Zhang, Yu, He, Wang, and Yi}]{jia2025omnispatialcomprehensivespatialreasoning}
Jia, M.; Qi, Z.; Zhang, S.; Zhang, W.; Yu, X.; He, J.; Wang, H.; and Yi, L. 2025.
\newblock OmniSpatial: Towards Comprehensive Spatial Reasoning Benchmark for Vision Language Models.
\newblock arXiv:2506.03135.

\bibitem[{Jin et~al.(2025)Jin, Yeom, Bae, and Kim}]{jin2025wellthinkingenhancingllm}
Jin, H.; Yeom, J.~W.; Bae, S.; and Kim, T. 2025.
\newblock "Well, Keep Thinking": Enhancing LLM Reasoning with Adaptive Injection Decoding.
\newblock arXiv:2503.10167.

\bibitem[{Kamath, Hessel, and Chang(2023)}]{kamath-etal-2023-whats}
Kamath, A.; Hessel, J.; and Chang, K.-W. 2023.
\newblock What{'}s ``up'' with vision-language models? Investigating their struggle with spatial reasoning.
\newblock In Bouamor, H.; Pino, J.; and Bali, K., eds., \emph{Proceedings of the 2023 Conference on Empirical Methods in Natural Language Processing}, 9161--9175. Singapore: Association for Computational Linguistics.

\bibitem[{Kolve et~al.(2017)Kolve, Mottaghi, Han, VanderBilt, Weihs, Herrasti, Deitke, Ehsani, Gordon, Zhu et~al.}]{kolve2017ai2}
Kolve, E.; Mottaghi, R.; Han, W.; VanderBilt, E.; Weihs, L.; Herrasti, A.; Deitke, M.; Ehsani, K.; Gordon, D.; Zhu, Y.; et~al. 2017.
\newblock Ai2-thor: An interactive 3d environment for visual ai.
\newblock \emph{arXiv preprint arXiv:1712.05474}.

\bibitem[{Krishna et~al.(2017)Krishna, Zhu, Groth, Johnson, Hata, Kravitz, Chen, Kalantidis, Li, Shamma et~al.}]{krishna2017visual}
Krishna, R.; Zhu, Y.; Groth, O.; Johnson, J.; Hata, K.; Kravitz, J.; Chen, S.; Kalantidis, Y.; Li, L.-J.; Shamma, D.~A.; et~al. 2017.
\newblock Visual genome: Connecting language and vision using crowdsourced dense image annotations.
\newblock \emph{International journal of computer vision}, 123(1): 32--73.

\bibitem[{Li et~al.(2024)Li, Zhang, Guo, Zhang, Li, Zhang, Zhang, Li, Liu, and Li}]{li2024llavaov}
Li, B.; Zhang, Y.; Guo, D.; Zhang, R.; Li, F.; Zhang, H.; Zhang, K.; Li, Y.; Liu, Z.; and Li, C. 2024.
\newblock Llava-onevision: Easy visual task transfer.
\newblock \emph{arXiv preprint arXiv:2408.03326}.

\bibitem[{Li et~al.(2025)Li, Wu, Zhang, Xia, Mao, Dong, Vulić, and Wei}]{li2025imaginereasoningspacemultimodal}
Li, C.; Wu, W.; Zhang, H.; Xia, Y.; Mao, S.; Dong, L.; Vulić, I.; and Wei, F. 2025.
\newblock Imagine while Reasoning in Space: Multimodal Visualization-of-Thought.
\newblock In \emph{Proceedings of the International Conference on Machine Learning}.

\bibitem[{Li et~al.(2023)Li, Zhang, Geng, Geng, Long, Shen, Zhang, Liu, and Dong}]{li2023manipllmembodiedmultimodallarge}
Li, X.; Zhang, M.; Geng, Y.; Geng, H.; Long, Y.; Shen, Y.; Zhang, R.; Liu, J.; and Dong, H. 2023.
\newblock ManipLLM: Embodied Multimodal Large Language Model for Object-Centric Robotic Manipulation.
\newblock arXiv:2312.16217.

\bibitem[{Liao et~al.(2025)Liao, Xie, Zhang, Kong, Lu, Yang, and Deng}]{liao2025improvedvisualspatialreasoningr1zerolike}
Liao, Z.; Xie, Q.; Zhang, Y.; Kong, Z.; Lu, H.; Yang, Z.; and Deng, Z. 2025.
\newblock Improved Visual-Spatial Reasoning via R1-Zero-Like Training.
\newblock arXiv:2504.00883.

\bibitem[{Lin et~al.(2025)Lin, Li, Zhong, Zou, and Zhao}]{lin2025evo}
Lin, T.; Li, G.; Zhong, Y.; Zou, Y.; and Zhao, B. 2025.
\newblock Evo-0: Vision-Language-Action Model with Implicit Spatial Understanding.
\newblock \emph{arXiv preprint arXiv:2507.00416}.

\bibitem[{Lin et~al.(2014)Lin, Maire, Belongie, Hays, Perona, Ramanan, Doll{\'a}r, and Zitnick}]{lin2014microsoft}
Lin, T.-Y.; Maire, M.; Belongie, S.; Hays, J.; Perona, P.; Ramanan, D.; Doll{\'a}r, P.; and Zitnick, C.~L. 2014.
\newblock Microsoft coco: Common objects in context.
\newblock In \emph{European conference on computer vision}, 740--755. Springer.

\bibitem[{Liu et~al.(2023)Liu, Li, Wu, and Lee}]{liu2023visual}
Liu, H.; Li, C.; Wu, Q.; and Lee, Y.~J. 2023.
\newblock Visual instruction tuning.
\newblock \emph{Advances in neural information processing systems}, 36: 34892--34916.

\bibitem[{Liu et~al.(2024)Liu, Yan, Zaharia, and Abbeel}]{liu2024world}
Liu, H.; Yan, W.; Zaharia, M.; and Abbeel, P. 2024.
\newblock World model on million-length video and language with ringattention.
\newblock \emph{arXiv preprint arXiv:2402.08268}.

\bibitem[{Liu et~al.(2025)Liu, Chi, Wu, Zhang, Hu, Zhang, Zhang, Wu, Cao, Huang, Huang, Tian, Qiu, Quan, Hao, and Zhuang}]{liu2025spatialcotadvancingspatialreasoning}
Liu, Y.; Chi, D.; Wu, S.; Zhang, Z.; Hu, Y.; Zhang, L.; Zhang, Y.; Wu, S.; Cao, T.; Huang, G.; Huang, H.; Tian, G.; Qiu, W.; Quan, X.; Hao, J.; and Zhuang, Y. 2025.
\newblock SpatialCoT: Advancing Spatial Reasoning through Coordinate Alignment and Chain-of-Thought for Embodied Task Planning.
\newblock arXiv:2501.10074.

\bibitem[{Luo et~al.(2025)Luo, Yang, Dou, Wang, Liu, Dai, Qiao, and Zhu}]{luo2025mono}
Luo, G.; Yang, X.; Dou, W.; Wang, Z.; Liu, J.; Dai, J.; Qiao, Y.; and Zhu, X. 2025.
\newblock Mono-internvl: Pushing the boundaries of monolithic multimodal large language models with endogenous visual pre-training.
\newblock In \emph{Proceedings of the Computer Vision and Pattern Recognition Conference}, 24960--24971.

\bibitem[{Ouyang et~al.(2025)Ouyang, Liu, Wu, Liu, Zhou, Zhou, Meng, and Sun}]{ouyang2025spacerreinforcingmllmsvideo}
Ouyang, K.; Liu, Y.; Wu, H.; Liu, Y.; Zhou, H.; Zhou, J.; Meng, F.; and Sun, X. 2025.
\newblock SpaceR: Reinforcing MLLMs in Video Spatial Reasoning.
\newblock arXiv:2504.01805.

\bibitem[{Radford et~al.(2021)Radford, Kim, Hallacy, Ramesh, Goh, Agarwal, Sastry, Askell, Mishkin, Clark, Krueger, and Sutskever}]{radford2021learningtransferablevisualmodels}
Radford, A.; Kim, J.~W.; Hallacy, C.; Ramesh, A.; Goh, G.; Agarwal, S.; Sastry, G.; Askell, A.; Mishkin, P.; Clark, J.; Krueger, G.; and Sutskever, I. 2021.
\newblock Learning Transferable Visual Models From Natural Language Supervision.
\newblock arXiv:2103.00020.

\bibitem[{Su et~al.(2024)Su, Ahmed, Lu, Pan, Bo, and Liu}]{su2024roformer}
Su, J.; Ahmed, M.; Lu, Y.; Pan, S.; Bo, W.; and Liu, Y. 2024.
\newblock Roformer: Enhanced transformer with rotary position embedding.
\newblock \emph{Neurocomputing}, 568: 127063.

\bibitem[{Tong et~al.(2024)Tong, Liu, Zhai, Ma, LeCun, and Xie}]{tong2024eyes}
Tong, S.; Liu, Z.; Zhai, Y.; Ma, Y.; LeCun, Y.; and Xie, S. 2024.
\newblock Eyes wide shut? exploring the visual shortcomings of multimodal llms.
\newblock In \emph{Proceedings of the IEEE/CVF Conference on Computer Vision and Pattern Recognition}, 9568--9578.

\bibitem[{Wei et~al.(2022)Wei, Wang, Schuurmans, Bosma, Xia, Chi, Le, Zhou et~al.}]{wei2022chain}
Wei, J.; Wang, X.; Schuurmans, D.; Bosma, M.; Xia, F.; Chi, E.; Le, Q.~V.; Zhou, D.; et~al. 2022.
\newblock Chain-of-thought prompting elicits reasoning in large language models.
\newblock \emph{Advances in neural information processing systems}, 35: 24824--24837.

\bibitem[{Wei et~al.(2025)Wei, Liu, Zang, Dong, Zhang, Cao, Tong, Duan, Guo, Wang, Qiu, and Lin}]{wei2025videoropemakesgoodvideo}
Wei, X.; Liu, X.; Zang, Y.; Dong, X.; Zhang, P.; Cao, Y.; Tong, J.; Duan, H.; Guo, Q.; Wang, J.; Qiu, X.; and Lin, D. 2025.
\newblock VideoRoPE: What Makes for Good Video Rotary Position Embedding?
\newblock arXiv:2502.05173.

\bibitem[{Wu et~al.(2025{\natexlab{a}})Wu, Liu, Hung, and Duan}]{wu2025spatialmllmboostingmllmcapabilities}
Wu, D.; Liu, F.; Hung, Y.-H.; and Duan, Y. 2025{\natexlab{a}}.
\newblock Spatial-MLLM: Boosting MLLM Capabilities in Visual-based Spatial Intelligence.
\newblock arXiv:2505.23747.

\bibitem[{Wu et~al.(2025{\natexlab{b}})Wu, Yang, Zhou, Fang, Song, Sun, and Ji}]{wu2025groundedchainofthoughtmultimodallarge}
Wu, Q.; Yang, X.; Zhou, Y.; Fang, C.; Song, B.; Sun, X.; and Ji, R. 2025{\natexlab{b}}.
\newblock Grounded Chain-of-Thought for Multimodal Large Language Models.
\newblock arXiv:2503.12799.

\bibitem[{Xiong et~al.(2024)Xiong, Shen, Li, Zhou, Liu, Wang, and Dong}]{xiong2024aicmllmautonomousinteractive}
Xiong, C.; Shen, C.; Li, X.; Zhou, K.; Liu, J.; Wang, R.; and Dong, H. 2024.
\newblock AIC MLLM: Autonomous Interactive Correction MLLM for Robust Robotic Manipulation.
\newblock In \emph{The Conference on Robot Learning}.

\bibitem[{Xu et~al.(2023)Xu, Xie, Qin, Tao, and Wang}]{xu2023parameter}
Xu, L.; Xie, H.; Qin, S.-Z.~J.; Tao, X.; and Wang, F.~L. 2023.
\newblock Parameter-efficient fine-tuning methods for pretrained language models: A critical review and assessment.
\newblock \emph{arXiv preprint arXiv:2312.12148}.

\bibitem[{Yang et~al.(2025)Yang, Yang, Gupta, Han, Fei-Fei, and Xie}]{yang2025thinkingspacemultimodallarge}
Yang, J.; Yang, S.; Gupta, A.~W.; Han, R.; Fei-Fei, L.; and Xie, S. 2025.
\newblock Thinking in Space: How Multimodal Large Language Models See, Remember, and Recall Spaces.
\newblock arXiv:2412.14171.

\bibitem[{Yeh et~al.(2025)Yeh, Wang, Tong, Cheng, Wang, Chu, Zhai, Chen, Gao, and Ma}]{yeh2025seeingperspectiveevaluatingmultiview}
Yeh, C.-H.; Wang, C.; Tong, S.; Cheng, T.-Y.; Wang, R.; Chu, T.; Zhai, Y.; Chen, Y.; Gao, S.; and Ma, Y. 2025.
\newblock Seeing from Another Perspective: Evaluating Multi-View Understanding in MLLMs.
\newblock In \emph{Neural Information Processing Systems}.

\bibitem[{Zhang et~al.(2025)Zhang, Khayatkhoei, Chhikara, and Ilievski}]{zhang2025mllms}
Zhang, J.; Khayatkhoei, M.; Chhikara, P.; and Ilievski, F. 2025.
\newblock {MLLM}s Know Where to Look: Training-free Perception of Small Visual Details with Multimodal {LLM}s.
\newblock In \emph{The Thirteenth International Conference on Learning Representations}.

\bibitem[{Zheng, Huang, and Wang(2025)}]{zheng2025video3dllmlearningpositionaware}
Zheng, D.; Huang, S.; and Wang, L. 2025.
\newblock Video-3D LLM: Learning Position-Aware Video Representation for 3D Scene Understanding.
\newblock arXiv:2412.00493.

\bibitem[{Zhou et~al.(2024)Zhou, He, Ke, Zhu, Gutierrez~Basulto, and Pan}]{zhou-etal-2024-empirical}
Zhou, X.; He, J.; Ke, Y.; Zhu, G.; Gutierrez~Basulto, V.; and Pan, J. 2024.
\newblock An Empirical Study on Parameter-Efficient Fine-Tuning for {M}ulti{M}odal Large Language Models.
\newblock In Ku, L.-W.; Martins, A.; and Srikumar, V., eds., \emph{Findings of the Association for Computational Linguistics}, 10057--10084. Bangkok, Thailand: Association for Computational Linguistics.

\bibitem[{Zhu et~al.(2025)Zhu, Wang, Chen, Liu, Ye, Gu, Tian, Duan, Su, Shao, Gao, Cui, Wang, Cao, Liu, Wei, Zhang, Wang, Xu, Li, Wang, Deng, Li, He, Jiang, Luo, Wang, He, Shi, Zhang, Shao, He, Xiong, Qu, Sun, Jiao, Lv, Wu, Zhang, Deng, Ge, Chen, Wang, Dou, Lu, Zhu, Lu, Lin, Qiao, Dai, and Wang}]{zhu2025internvl3exploringadvancedtraining}
Zhu, J.; Wang, W.; Chen, Z.; Liu, Z.; Ye, S.; Gu, L.; Tian, H.; Duan, Y.; Su, W.; Shao, J.; Gao, Z.; Cui, E.; Wang, X.; Cao, Y.; Liu, Y.; Wei, X.; Zhang, H.; Wang, H.; Xu, W.; Li, H.; Wang, J.; Deng, N.; Li, S.; He, Y.; Jiang, T.; Luo, J.; Wang, Y.; He, C.; Shi, B.; Zhang, X.; Shao, W.; He, J.; Xiong, Y.; Qu, W.; Sun, P.; Jiao, P.; Lv, H.; Wu, L.; Zhang, K.; Deng, H.; Ge, J.; Chen, K.; Wang, L.; Dou, M.; Lu, L.; Zhu, X.; Lu, T.; Lin, D.; Qiao, Y.; Dai, J.; and Wang, W. 2025.
\newblock InternVL3: Exploring Advanced Training and Test-Time Recipes for Open-Source Multimodal Models.
\newblock arXiv:2504.10479.

\end{thebibliography}

% ----------- Supplementary Content Starts Here -----------
\newpage
\appendix

\section{Appendices}

\subsection{Performance Analysis on VSI-bench}
\label{app:exp_vsi}
 
This section provides a detailed breakdown of the model's performance across all sub-tasks in the VSI-Bench under increasing training data scales. As detailed in the main paper, our model was fine-tuned on subsets of the synthetic SR-91k dataset, with training sizes ranging from 10k to 90k samples, before being evaluated on VSI-Bench.
 
The comprehensive results are tabulated in Table~\ref{tab:data_train_vsi_all}. A consistent trend observed across all sub-tasks is a general performance improvement with an increasing amount of training data. However, the magnitude of this improvement is highly task-dependent. For instance, tasks that demand a higher level of abstract reasoning, such as Route Plan task, exhibit limited performance gains even with a significantly larger training set. Conversely, tasks that primarily rely on spatial-semantic matching, like Appearance Order task, demonstrate rapid and substantial improvements. 
 
This observed divergence between task types further substantiates our data-centriv analysis, highlighting that the performance ceiling is intrinsically linked to the nature of the task. While simply scaling up the training data volume provides diminishing returns for enhancing MLLMs' spatial understanding capabilities, the upper bound is relatively low, especially for tasks that require spatial imagination.

\begin{table*}[t]
\centering
% \small
\caption{Performance under increasing training data scales on Qwen2.5-VL-7B for video-based scenarios.}
\begin{tabular}{ccccccccccc}
\toprule
Data Size 
& Appr. Order 
& Abs. Dist. 
& Counting 
& Rel. Dist. 
& Obj. Size 
& Room Size 
& Route Plan 
& Rel. Dir.
& Overall \\
\midrule
0 & 29.60 & 21.20 & 38.60 & 37.30 & 49.30 & 37.10 & 30.00 & 39.50 & 35.40 \\
10k    & 47.70 & 26.20 & 53.70 & 41.50 & 59.30 & 48.60 & 31.40 & 45.90 & 44.30 \\
20k    & 49.80 & 26.40 & 54.00 & 41.60 & 58.30 & 48.40 & 32.90 & 45.40 & 44.60 \\
30k    & 53.00 & 28.90 & 55.50 & 41.90 & 60.50 & 47.90 & 31.40 & 45.60 & 45.60 \\
40k    & 55.50 & 29.50 & 56.80 & 42.80 & 62.10 & 48.90 & 30.40 & 46.30 & 46.50 \\
50k    & 55.30 & 30.40 & 59.20 & 43.00 & 61.60 & 50.70 & 31.90 & 46.00 & 47.30 \\
60k    & 55.10 & 27.90 & 57.60 & 44.30 & 62.60 & 47.50 & 26.80 & 46.30 & 46.00 \\
70k    & 52.70 & 29.20 & 58.70 & 44.50 & 64.40 & 48.10 & 29.80 & 46.90 & 46.80 \\
80k    & 53.50 & 28.80 & 59.10 & 45.90 & 63.80 & 48.20 & 28.30 & 45.90 & 46.70 \\
90k    & 53.80 & 31.20 & 63.20 & 44.30 & 62.70 & 47.70 & 31.90 & 46.50 & 47.70 \\
\bottomrule
\end{tabular}
\label{tab:data_train_vsi_all}
\end{table*}

\subsection{Detailed Ablation Study on Positional Encodings}
\label{app:ablation_details}
 
This section presents the comprehensive results of our ablation study on positional encodings (PEs), providing a granular view that complements the analysis in our main paper. We evaluate the impact of various PE ablation strategies(Mask, Shuffle, and Constant Value), on the performance of Qwen2.5-VL-7B and LLaVA-OV-7B across all single-view and multi-view spatial reasoning tasks.
 
The detailed outcomes for single-view tasks are presented in Table~\ref{tb:merged_rope_comparison_qwen} for Qwen2.5-VL-7B and Table~\ref{tb:merged_rope_comparison_llava} for LLaVA-OV-7B. The results for multi-view tasks for both models are consolidated in Table~\ref{tb:rope_comparison_multi-view}. Collectively, these tables provide robust empirical evidence for the conclusions drawn in our main analysis.
 
Our findings consistently demonstrate that the positional encoding from the Vision Encoder (VE) is paramount for spatial understanding. As shown across all three tables, ablating the VE's RoPE (Mask-VE, Shuffle-VE) results in a catastrophic performance decline across nearly all spatial tasks. This effect is particularly pronounced in the cascaded model LLaVA-OneVision(Table~\ref{tb:merged_rope_comparison_llava}), which exhibits a greater dependency on explicit VE positional signals compared to the more integrated model Qwen2.5-VL-7B(Table~\ref{tb:merged_rope_comparison_qwen}). This corroborates our claim that model architecture influences the reliance on different sources of positional information.

\begin{table*}[h]
\centering
\small
\caption{Performance of different RoPE ablations for Qwen2.5-VL-7B on all single-view sub-tasks.}
\setlength\tabcolsep{3pt}
\begin{tabular}{lccccccccc}
\toprule
\textbf{Ablation Type} 
& \makecell{\textbf{Controlled} \\ \textbf{Images\_A}} 
& \makecell{\textbf{Controlled} \\ \textbf{Images\_B}} 
& \makecell{\textbf{COCO\_QA} \\ \textbf{one\_obj}} 
& \makecell{\textbf{COCO\_QA} \\ \textbf{two\_obj}} 
& \makecell{\textbf{VG\_QA} \\ \textbf{one\_obj}} 
& \makecell{\textbf{VG\_QA} \\ \textbf{two\_obj}} 
& \makecell{\textbf{VG} \\ \textbf{Relation}}  
& \makecell{\textbf{VG} \\ \textbf{Attribution}}  \\
\midrule
vanilla             & 79.42 & 93.89 & 88.26 & 86.39 & 91.30 & 88.70 & 91.59 & 90.79 \\
\midrule
Mask-LLM-txy(*)     & 0.00(-79.42)  & 0.73(-93.16)  & 0.22(-88.04)  & 0.00(-86.39)  & 0.43(-90.87)  & 0.00(-88.70)  & 0.06(-91.53)  & 0.05(-90.74)  \\
Mask-LLM-xy(*)      & 82.08(+2.66)  & 87.78(-6.11)  & 79.80(-8.46)  & 81.18(-5.21)  & 89.23(-2.07)  & 88.01(-0.69)  & 90.52(-1.07)  & 90.63(-0.16)  \\
Mask-LLM-x(*)       & 81.84(+2.42)  & 89.00(-4.89)  & 80.20(-8.06)  & 81.18(-5.21)  & 88.89(-2.41)  & 87.67(-1.03)  & 90.47(-1.12)  & 90.64(-0.15)  \\
Mask-LLM-y(*)       & 79.66(+0.24)  & 93.64(-0.25)  & 88.17(-0.09)  & 86.85(+0.46)  & 90.96(-0.34)  & 89.38(+0.68)  & 91.70(+0.11)  & 90.90(+0.11)  \\
Shuffle-LLM-txy(*)  & 3.15(-76.27)  & 2.69(-91.20)  & 1.02(-87.24)  & 0.90(-85.49)  & 1.64(-89.66)  & 1.37(-87.33)  & 1.51(-90.08)  & 1.63(-89.16)  \\
Shuffle-LLM-xy(*)   & 78.69(-0.73)  & 82.89(-11.00) & 77.36(-10.90) & 76.19(-10.20) & 87.77(-3.53)  & 82.53(-6.17)  & 88.99(-2.60)  & 90.61(-0.18)  \\
Shuffle-LLM-x(*)    & 79.42(+0.00)  & 81.17(-12.72) & 79.18(-9.08)  & 76.42(-9.97)  & 88.11(-3.19)  & 83.22(-5.48)  & 89.05(-2.54)  & 90.84(+0.05)  \\
Shuffle-LLM-y(*)    & 79.18(-0.24)  & 93.89(+0.00)  & 87.94(-0.32)  & 87.07(+0.68)  & 90.96(-0.34)  & 89.38(+0.68)  & 91.67(+0.08)  & 90.85(+0.06)  \\
\midrule
Mask-LLM-txy        & 88.14(+8.72)  & 92.91(-0.98)  & 86.52(-1.74)  & 86.39(+0.00)  & 90.53(-0.77)  & 89.38(+0.68)  & 90.74(-0.85)  & 89.45(-1.34)  \\
Mask-LLM-xy         & 86.44(+7.02)  & 92.67(-1.22)  & 85.72(-2.54)  & 87.07(+0.68)  & 89.75(-1.55)  & 89.04(+0.34)  & 90.97(-0.62)  & 90.59(-0.20)  \\
Mask-LLM-x          & 86.44(+7.02)  & 92.67(-1.22)  & 85.72(-2.54)  & 87.07(+0.68)  & 89.75(-1.55)  & 89.04(+0.34)  & 90.97(-0.62)  & 90.59(-0.20)  \\
Mask-LLM-y          & 79.90(+0.48)  & 93.89(+0.00)  & 87.94(-0.32)  & 87.07(+0.68)  & 91.13(-0.17)  & 88.70(+0.00)  & 91.57(-0.02)  & 90.83(+0.04)  \\
Shuffle-LLM-txy     & 76.51(-2.91)  & 85.82(-8.07)  & 80.03(-8.23)  & 82.77(-3.62)  & 86.74(-4.56)  & 89.04(+0.34)  & 88.36(-3.23)  & 87.92(-2.87)  \\
Shuffle-LLM-xy      & 85.47(+6.05)  & 93.50(-0.39)  & 84.56(-3.70)  & 85.71(-0.68)  & 89.66(-1.64)  & 88.36(-0.34)  & 91.29(-0.30)  & 90.62(-0.17)  \\
Shuffle-LLM-x       & 84.50(+5.08)  & 93.15(-0.74)  & 84.70(-3.56)  & 95.49(+9.10)  & 89.41(-1.89)  & 88.36(-0.34)  & 91.32(-0.27)  & 90.70(-0.09)  \\
Shuffle-LLM-y       & 79.66(+0.24)  & 93.89(+0.00)  & 88.03(-0.23)  & 86.17(-0.22)  & 91.13(-0.17)  & 88.36(-0.34)  & 91.62(+0.03)  & 90.80(+0.01)  \\
\midrule
Mask-VE-hw          & 11.38(-68.04) & 6.60(-87.29)  & 44.93(-43.33) & 24.94(-61.45) & 36.95(-54.35) & 16.78(-71.92) & 81.63(-9.96)  & 85.85(-4.94)  \\
Mask-VE-h           & 46.49(-32.93) & 52.08(-41.81) & 67.79(-20.47) & 55.10(-31.29) & 68.56(-22.74) & 61.99(-26.71) & 91.48(-0.11)  & 87.13(-3.66)  \\
Mask-VE-w           & 23.00(-56.42) & 19.07(-74.82) & 53.87(-34.39) & 32.88(-53.51) & 45.22(-46.08) & 28.08(-60.62) & 80.13(-11.46) & 89.08(-1.71)  \\
Shuffle-VE-hw       & 8.47(-70.95)  & 11.00(-82.89) & 49.60(-38.66) & 38.32(-48.07) & 47.29(-44.01) & 26.37(-62.33) & 79.05(-12.54) & 82.81(-7.98)  \\
Shuffle-VE-h        & 26.63(-52.79) & 25.43(-68.46) & 72.33(-15.93) & 60.09(-26.30) & 69.77(-21.53) & 60.62(-28.08) & 90.78(-0.81)  & 84.63(-6.16)  \\
Shuffle-VE-w        & 12.83(-66.59) & 17.60(-76.29) & 50.44(-37.82) & 29.93(-56.46) & 43.41(-47.89) & 19.52(-69.18) & 81.26(-10.33) & 86.74(-4.05)  \\
\bottomrule
\end{tabular}
\label{tb:merged_rope_comparison_qwen}
\end{table*}

\begin{table*}[h]
\centering
\small
\setlength\tabcolsep{3pt}
\caption{Performance of different RoPE ablations for Llava-OV-7B on all single-view sub-tasks.}
\begin{tabular}{lcccccccc}
\toprule
\textbf{Ablation Type} 
& \makecell{\textbf{Controlled} \\ \textbf{Images\_A}} 
& \makecell{\textbf{Controlled} \\ \textbf{Images\_B}} 
& \makecell{\textbf{COCO\_QA} \\ \textbf{one\_obj}} 
& \makecell{\textbf{COCO\_QA} \\ \textbf{two\_obj}} 
& \makecell{\textbf{VG\_QA} \\ \textbf{one\_obj}} 
& \makecell{\textbf{VG\_QA} \\ \textbf{two\_obj}} 
& \makecell{\textbf{VG} \\ \textbf{Relation}}  
& \makecell{\textbf{VG} \\ \textbf{Attribution}}  \\
\midrule
vanilla & 98.79 & 98.53 & 93.90 & 79.32 & 97.59 & 95.88 & 90.55 & 94.33 \\
\midrule
Mask-LLM-txy(*) & 0.00(-98.79) & 0.00(-98.53) & 0.22(-93.68) & 0.00(-79.32) & 0.26(-97.33) & 0.00(-95.88) & 0.00(-90.55) & 0.00(-94.33) \\
Shuffle-LLM-txy(*) & 0.00(-98.79) & 0.00(-98.53) & 0.00(-93.90) & 0.00(-79.32) & 0.00(-97.59) & 0.00(-95.88) & 0.00(-90.55) & 0.00(-94.33) \\
\midrule
Const-LLM-First & 97.09(-1.70) & 98.53(+0.00) & 72.81(-21.09) & 66.14(-13.18) & 78.45(-19.14) & 82.47(-13.41) & 83.62(-6.93) & 96.42(+2.09) \\
Const-LLM-Last & 0.00(-98.79) & 0.00(-98.53) & 53.49(-40.41) & 30.23(-49.09) & 69.40(-28.19) & 58.76(-37.12) & 75.70(-14.85) & 85.54(-8.79) \\
Shuffle-LLM-txy & 98.54(-0.25) & 96.81(-1.72) & 78.91(-14.99) & 66.14(-13.18) & 85.95(-11.64) & 85.91(-9.97) & 84.74(-5.81) & 94.27(-0.06) \\
\midrule
Const-VE-First & 17.23(-81.56) & 4.41(-94.12) & 0.09(-93.81) & 5.45(-73.87) & 1.38(-96.21) & 6.53(-89.35) & 77.21(-13.34) & 98.77(+4.44) \\
Const-VE-Last & 18.20(-80.59) & 3.43(-95.10) & 0.13(-93.77) & 5.45(-73.87) & 0.95(-96.64) & 6.53(-89.35) & 78.53(-12.02) & 98.74(+4.41) \\
Shuffle-VE-hw & 14.56(-84.23) & 10.05(-88.48) & 36.45(-57.45) & 11.36(-67.96) & 38.45(-59.14) & 12.71(-83.17) & 95.71(+5.16) & 95.84(+1.51) \\
Shuffle-VE-h & 79.13(-19.66) & 68.63(-29.90) & 82.96(-10.94) & 66.14(-13.18) & 84.14(-13.45) & 82.82(-13.06) & 91.39(+0.84) & 92.78(-1.55) \\
Shuffle-VE-w & 68.20(-30.59) & 64.46(-34.07) & 76.64(-17.26) & 62.50(-16.82) & 79.14(-18.45) & 77.32(-18.56) & 88.43(-2.12) & 93.04(-1.29) \\
\bottomrule
\end{tabular}
\label{tb:merged_rope_comparison_llava}
\end{table*}

%=============================================================

\begin{table*}[h]
\centering
\setlength\tabcolsep{6pt}
\caption{Performance of Qwen2.5-VL-7B and Llava-OV-7B under different Ablation Type on MulSeT.}
\begin{tabular}{llcccc}
\toprule
\textbf{Model} & \textbf{Ablation Type} & 
\makecell{\textbf{Occlusion} \\ \textbf{Restoration}} & 
\makecell{\textbf{Distance} \\ \textbf{Comparison}} & 
\makecell{\textbf{Azimuth} \\ \textbf{Transfer}} & 
\textbf{Overall} \\
\midrule

\multirow{22}{*}{Qwen2.5-VL-7b} 
& Vanilla         & 34.52 & 43.41 & 27.25 & 35.06\\
& Mask-LLM-txy(*) & 0.00 (-34.52) & 0.00 (-43.41) & 0.00 (-27.25) & 0.00(-35.06) \\
& Mask-LLM-xy(*)  & 30.84 (-3.68) & 41.70 (-1.71) & 29.76 (+2.51) & 34.10(-0.96) \\
& Mask-LLM-x(*)   & 31.54 (-2.98) & 41.84 (-1.57) & 30.99 (+3.74) & 34.79(-0.27) \\
& Mask-LLM-y(*)   & 34.93 (+0.41) & 43.21 (-0.20) & 28.42 (+1.17) & 35.52(+0.46) \\
& Shuffle-LLM-txy(*) & 0.41 (-34.11) & 0.85 (-42.56) & 0.61 (-26.64) & 0.62(-34.44) \\
& Shuffle-LLM-xy(*)  & 26.64 (-7.88) & 38.84 (-4.57) & 30.54 (+3.29) & 32.01(-3.05) \\
& Shuffle-LLM-x(*)   & 25.86 (-8.66) & 39.52 (-3.89) & 29.59 (+2.34) & 31.66(-3.40) \\
& Shuffle-LLM-y(*)   & 34.81 (+0.29) & 43.07 (-0.34) & 28.20 (+0.95) & 35.36(+0.30) \\
& Mask-LLM-txy  & 24.96 (-9.56) & 42.66 (-0.75) & 29.54 (+2.29) & 32.39(-2.67) \\
& Mask-LLM-xy   & 34.03 (-0.49) & 42.70 (-0.71) & 28.36 (+1.11) & 35.03(-0.03) \\
& Mask-LLM-x    & 34.00 (-0.52) & 42.53 (-0.88) & 27.36 (+0.11) & 34.63(-0.43) \\
& Mask-LLM-y    & 34.61 (+0.09) & 43.48 (+0.07) & 27.41 (+0.16) & 35.17(+0.11) \\
& Shuffle-LLM-txy & 37.19 (+2.67) & 39.01 (-4.40) & 33.78 (+6.53) & 36.66(+1.60) \\
& Shuffle-LLM-xy & 33.83 (-0.69) & 43.14 (-0.27) & 27.86 (+0.61) & 34.94(-0.12) \\
& Shuffle-LLM-x  & 34.55 (+0.03) & 42.63 (-0.78) & 27.41 (+0.16) & 34.86(-0.20) \\
& Shuffle-LLM-y  & 34.75 (+0.23) & 43.48 (+0.07) & 27.47 (+0.22) & 35.23(+0.17) \\
& Shuffle-VE-hw  & 22.47 (-12.05) & 36.89 (-6.52) & 30.71 (+3.46) & 30.02(-5.04) \\
& Shuffle-VE-h   & 30.35 (-4.17) & 36.86 (-6.55) & 33.05 (+5.80) & 33.42(-1.97) \\
& Shuffle-VE-w   & 28.38 (-6.14) & 36.69 (-6.72) & 33.17 (+5.92) & 32.75(-2.31) \\
& Mask-VE-hw     & 25.31 (-9.21) & 26.35 (-17.06) & 34.56 (+7.31) & 28.74(-6.32) \\
& Mask-VE-h      & 25.36 (-9.16) & 33.48 (-9.93) & 33.28 (+6.03) & 30.71(-4.35) \\
& Mask-VE-w      & 26.44 (-8.08) & 27.06 (-16.35) & 34.84 (+7.59) & 29.45(-5.61) \\

\midrule
\multirow{12}{*}{Llava-OV-7b} 
& Vanilla               & 22.38 & 43.75 & 24.67 & 30.27\\
& Shuffle-LLM-txy       & 18.82 (-3.56) & 41.94 (-1.81) & 22.50 (-2.17) & 27.75(-2.51) \\
& Const-LLM-First       & 19.80 (-2.58) & 40.10 (-3.65) & 23.05 (-1.62) & 27.65(-2.62) \\
& Const-LLM-Last        & 22.35 (-0.03) & 35.52 (-8.23) & 21.88 (-2.79) & 26.58(-3.68) \\
& Shuffle-VE-hw         & 16.36 (-6.02) & 35.11 (-8.64) & 22.44 (-2.23) & 24.64(-5.63) \\
& VE-noise              & 13.26 (-9.12) & 37.20 (-6.55) & 21.10 (-3.57) & 23.85(-6.41) \\
& Const-LLM(*)          & 0.00 (-22.38) & 0.00 (-43.75) & 0.00 (-24.67) & 0.00(-30.27) \\
& Shuffle-LLM(*)        & 0.00 (-22.38) & 0.00 (-43.75) & 0.00 (-24.67) & 0.00(-30.27) \\
& Const-VE-First        & 11.96 (-10.42) & 35.70 (-8.05) & 20.44 (-4.23) & 22.70(-7.57) \\
& Const-VE-Last         & 11.99 (-10.39) & 35.67 (-8.08) & 20.49 (-4.18) & 22.72(-7.55) \\
& Mask-VE-h             & 18.48 (-3.90) & 37.30 (-6.45) & 24.29 (-0.38) & 26.69(-3.58) \\
& Mask-VE-w             & 17.55 (-4.83) & 38.67 (-5.08) & 22.50 (-2.17) & 26.24(-4.03) \\

\bottomrule
\end{tabular}
\label{tb:rope_comparison_multi-view}
\end{table*}

\subsection{Directional Ablation Experiment}
\label{app:directional_ablation}
 
To further investigate the functional roles of the Vision Encoder's (VE) positional encodings, we conducted a directional ablation study. This analysis, referenced in the main text, aims to verify that the height ($h$) and width ($w$) components of the VE's RoPE directly correlate with the model's ability to discriminate along vertical and horizontal spatial axes, respectively. For this purpose, we evaluated the Qwen2.5-VL-7B and LLaVA-OV-7B models on two specialized datasets, Controlled\_Images\_A and Controlled\_Images\_B, which are explicitly designed to test performance on directional spatial judgments (e.g., left/right, up/down, and front/back).
 
The results are presented in Table~\ref{tab:rope_direction_qwen} for Qwen2.5-VL-7B and Table~\ref{tab:rope_direction_llava} for LLaVA-OV-7B. The findings reveal a clear and strong correlation between the ablated PE dimension and the corresponding spatial discrimination task. 
 
Specifically, ablating the width ($w$) dimension of the VE's RoPE (e.g., via Mask-VE-w) causes a catastrophic drop in performance on tasks requiring left-right discrimination. The impact of this same ablation on up-down or front-back judgments is comparatively minor. Conversely, ablating the height ($h$) dimension (e.g., via Mask-VE-h) severely impairs the model's ability to discern vertical or depth-related spatial relationships, while having a much smaller effect on horizontal judgments.
 
This strong, disentangled relationship provides compelling evidence that the VE's positional encodings are functionally specialized. The `w` and `h` dimensions are not merely abstract positional markers; they directly encode the horizontal and vertical arrangement of visual features, respectively. This confirms the mechanism hypothesized in our main analysis and underscores the importance of these explicit spatial signals for foundational spatial understanding.

\begin{table*}[t]
\centering
% \small
\setlength\tabcolsep{7pt}
\caption{Directional ablation results for Qwen2.5-VL-7B on the Controlled\_Images datasets. }
\begin{tabular}{lrrrrrrrrrr}
\toprule
\textbf{Ablation Type} 
& \multicolumn{5}{c}{\textbf{Controlled\_Images\_A}} 
& \multicolumn{5}{c}{\textbf{Controlled\_Images\_B}} \\
\cmidrule(lr){2-6} \cmidrule(lr){7-11}
& left & right & on & under & total 
& left & right & front & behind & total \\
\midrule
vanilla               & 43.69 & 91.26 & 100.00 & 83.50 & 79.42 & 100.00 & 98.04 & 78.43 & 100.00 & 93.89 \\
Mask-LLM-txy (*)         & 0.00 & 0.00 & 0.00 & 0.00 & 0.00 & 0.00 & 0.00 & 1.96 & 0.98 & 0.73 \\
Mask-LLM-xy (*)          & 60.19 & 89.32 & 99.03 & 80.58 & 82.08 & 99.02 & 95.10 & 65.69 & 92.16 & 87.78 \\
Mask-LLM-x (*)            & 60.19 & 90.29 & 98.06 & 79.61 & 81.84 & 99.02 & 97.06 & 67.65 & 93.14 & 89.00 \\
Mask-LLM-y (*)           & 45.63 & 91.26 & 100.00 & 82.52 & 79.66 & 100.00 & 98.04 & 77.45 & 100.00 & 93.64 \\
Shuffle-LLM-txy (*)        & 2.91 & 2.91 & 2.91 & 3.88 & 3.15 & 2.94 & 4.90 & 0.98 & 1.96 & 2.69 \\
Shuffle-LLM-xy (*)         & 58.25 & 86.41 & 95.15 & 75.73 & 78.69 & 95.10 & 88.24 & 57.84 & 91.18 & 82.89 \\
Shuffle-LLM-x (*)          & 60.19 & 82.52 & 95.15 & 80.58 & 79.42 & 89.22 & 91.18 & 54.90 & 90.20 & 81.17 \\
Shuffle-LLM-y (*)          & 43.69 & 91.26 & 100.00 & 82.52 & 79.18 & 100.00 & 98.04 & 78.43 & 100.00 & 93.89 \\
\midrule
Mask-LLM-txy        & 73.79 & 93.20 & 99.03 & 87.38 & 88.14 & 100.00 & 99.02 & 75.49 & 98.04 & 92.91 \\
Mask-LLM-xy         & 68.93 & 94.17 & 98.06 & 85.44 & 86.44 & 100.00 & 99.02 & 73.53 & 99.02 & 92.67 \\
Mask-LLM-x & 68.93 & 94.17 & 98.06 & 85.44 & 86.44 & 100.00 & 99.02 & 73.53 & 99.02 & 92.67 \\
Mask-LLM-y          & 44.66 & 91.26 & 100.00 & 84.47 & 79.90 & 100.00 & 98.04 & 78.43 & 100.00 & 93.89 \\
Shuffle-LLM-txy      & 40.78 & 92.23 & 99.03 & 74.76 & 76.51 & 95.10 & 95.10 & 54.90 & 99.02 & 85.82 \\
Shuffle-LLM-xy       & 64.08 & 94.17 & 99.03 & 85.44 & 85.47 & 100.00 & 99.02 & 75.49 & 100.00 & 93.50 \\
Shuffle-LLM-x        & 61.17 & 94.17 & 98.06 & 85.44 & 84.50 & 99.02 & 97.06 & 77.45 & 100.00 & 93.15 \\
Shuffle-LLM-y        & 44.66 & 91.26 & 100.00 & 83.50 & 79.66 & 100.00 & 98.04 & 78.43 & 100.00 & 93.89 \\
\midrule
Mask-VE-hw        & 0.00 & 0.00 & 0.00 & 45.63 & 11.38 & 2.94 & 0.00 & 4.90 & 18.63 & 6.60 \\
Mask-VE-h         & 46.60 & 78.64 & 32.04 & 29.13 & 46.49 & 50.98 & 65.69 & 29.41 & 62.75 & 52.08 \\
Mask-VE-w         & 0.97 & 0.97 & 35.92 & 54.37 & 23.00 & 7.84 & 2.94 & 26.47 & 39.22 & 19.07 \\
Shuffle-VE-hw      & 0.00 & 0.00 & 0.00 & 33.98 & 8.47 & 0.00 & 0.00 & 6.86 & 37.25 & 11.00 \\
Shuffle-VE-h       & 13.59 & 52.43 & 16.50 & 24.27 & 26.63 & 6.86 & 17.65 & 13.73 & 63.73 & 25.43 \\
Shuffle-VE-w       & 0.00 & 0.00 & 12.62 & 38.83 & 12.83 & 2.94 & 0.98 & 15.69 & 50.98 & 17.60 \\
\bottomrule
\end{tabular}
\label{tab:rope_direction_qwen}
\end{table*}

%=============================================================
\begin{table*}[t]
\centering
% \resizebox{\linewidth}{!}{%
\setlength\tabcolsep{7pt}
\renewcommand{\arraystretch}{1.2}
\caption{
Directional ablation results for Llava-OV-7B on the Controlled\_Images datasets. 
}
\begin{tabular}{lccccc ccccc}
\toprule
\textbf{Ablation Type}
& \multicolumn{5}{c}{\textbf{Controlled\_Images\_A}} 
& \multicolumn{5}{c}{\textbf{Controlled\_Images\_B}} \\
\cmidrule(lr){2-6} \cmidrule(lr){7-11}
& left & right & on & under & total
& left & right & front & behind & total \\
\midrule
Shuffle-VE-h & 0.8350 & 0.8641 & 0.7573 & 0.7087 & 0.7913 & 0.9510 & 0.9510 & 0.4412 & 0.4020 & 0.6863 \\
Shuffle-VE-w & 0.4854 & 0.4757 & 0.9223 & 0.8447 & 0.6820 & 0.5196 & 0.5882 & 0.6863 & 0.7843 & 0.6446 \\
\bottomrule
\end{tabular}
% }
\label{tab:rope_direction_llava}
\end{table*}
%=============================================================

%====================================================================
\begin{figure}[h]
\centering
\includegraphics[width=1\linewidth]{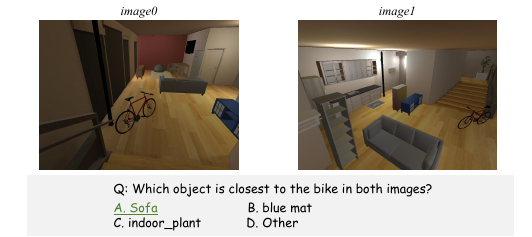}
\caption{Case study of attention visualization. }
\label{fig:attention_qa}
\end{figure}
%====================================================================

\subsection{Comprehensive Attention Visualization}
\label{app:attention_viz}
 
This section provides a comprehensive, layer-by-layer visualization of the model's attention maps under different inference methods, supplementing the targeted analysis presented in the main body of the paper. While the main text highlights the most illustrative layers (e.g., 21st and 27th), here we present the complete attention progression from layer 1 to 28. The visualizations correspond to the question-image pair shown in Figure~\ref{fig:attention_qa}. 
 
\begin{figure*}[h]
\centering
\includegraphics[width=1\linewidth]{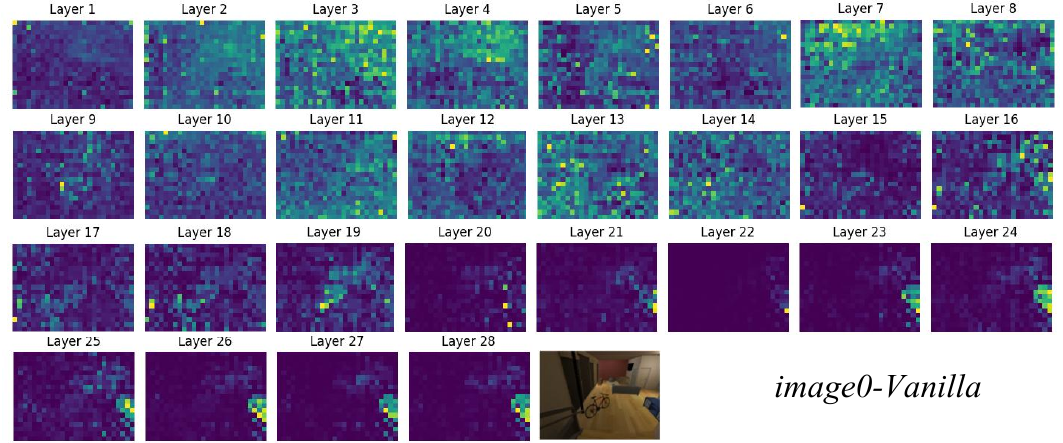}
\caption{Layer-wise attention visualization of image0 in Qwen2.5-VL-7B for the case input with the vanilla prompt.}
\label{fig:attention_image0-Vanilla}
\end{figure*}

\begin{figure*}[h]
\centering
\includegraphics[width=1\linewidth]{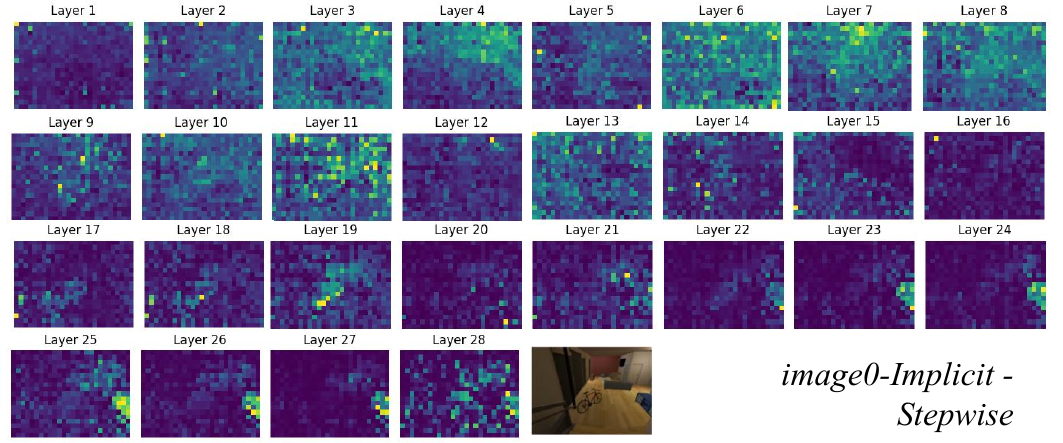}
\caption{Layer-wise attention visualization of image0 in Qwen2.5-VL-7B for the case input with the Implicit Stepwise prompt.}
\label{fig:attention_image0-Implicit-Stepwise}
\end{figure*}

\begin{figure*}[h]
\centering
\includegraphics[width=1\linewidth]{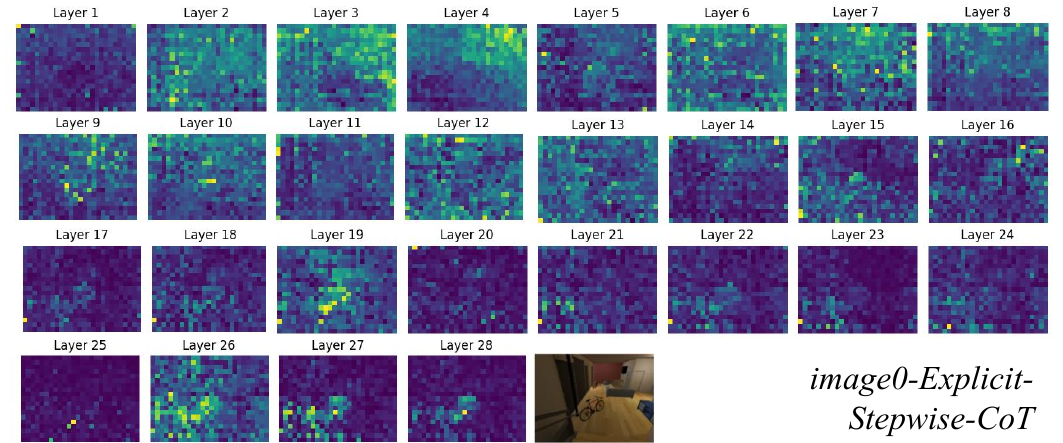}
\caption{Layer-wise attention visualization of image0 in Qwen2.5-VL-7B for the case input with the Explicit Stepwise CoT prompt.}
\label{fig:attention_image0-Explicit-Stepwise-CoT}
\end{figure*}

\begin{figure*}[h]
\centering
\includegraphics[width=1\linewidth]{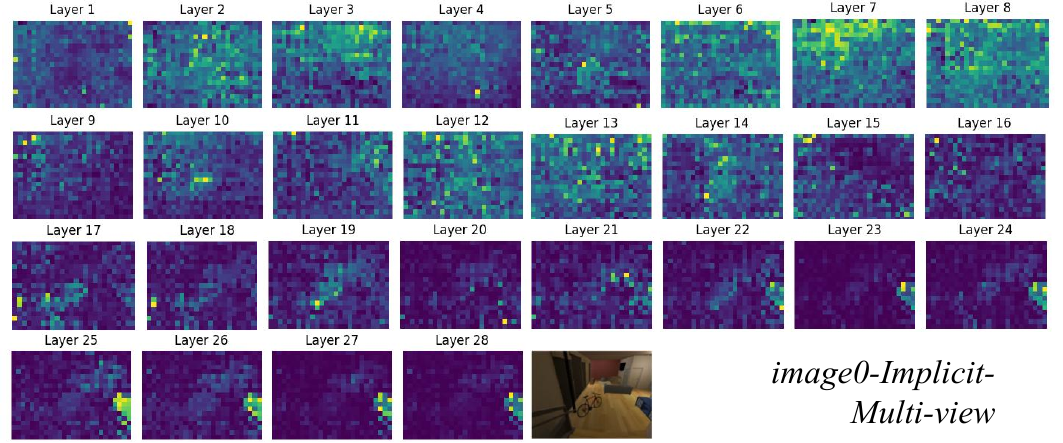}
\caption{Layer-wise attention visualization of image0 in Qwen2.5-VL-7B for the case input with the Implicit Multi-view prompt.}
\label{fig:attention_image0-Implicit-Multi-view}
\end{figure*}

\begin{figure*}[h]
\centering
\includegraphics[width=1\linewidth]{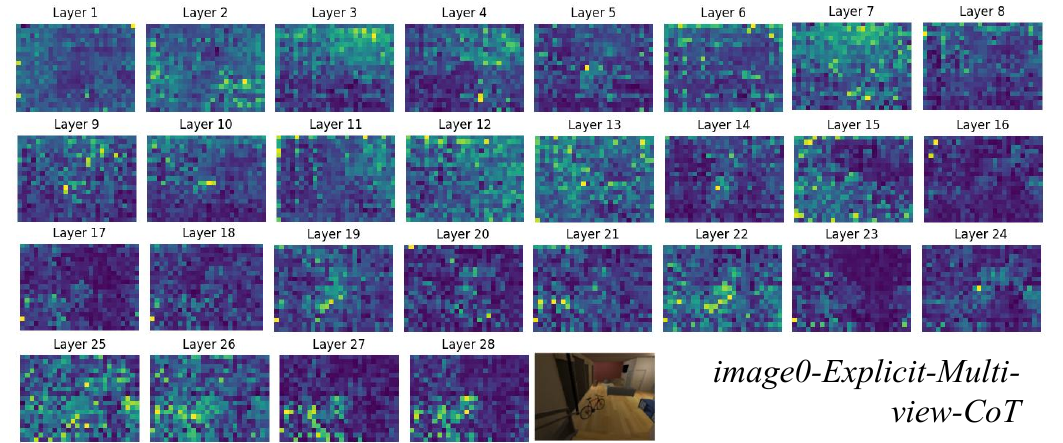}
\caption{Layer-wise attention visualization of image0 in Qwen2.5-VL-7B for the case input with the Explicit Multi-view CoT prompt.}
\label{fig:attention_image0-Explicit-Multi-view-CoT}
\end{figure*}

\begin{figure*}[h]
\centering
\includegraphics[width=1\linewidth]{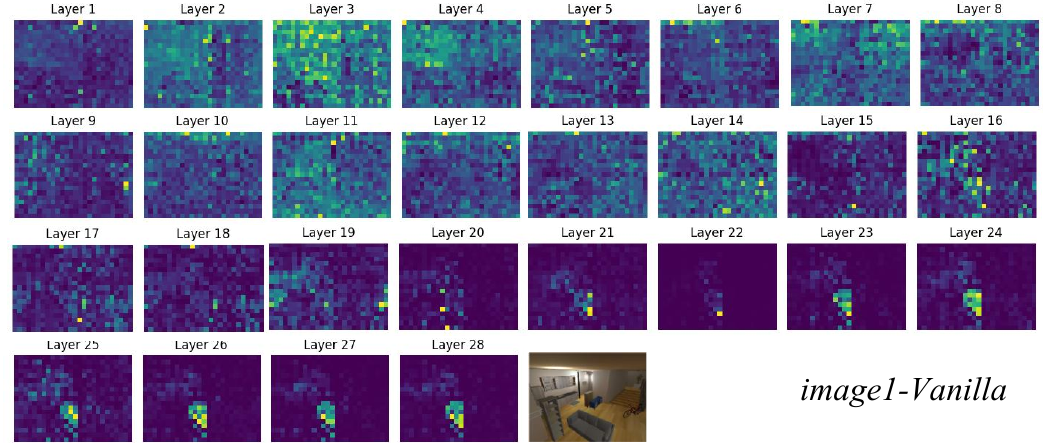}
\caption{Layer-wise attention visualization of image1 in Qwen2.5-VL-7B for the case input with the vanilla prompt.}
\label{fig:attention_image1-Vanilla}
\end{figure*}

\begin{figure*}[h]
\centering
\includegraphics[width=1\linewidth]{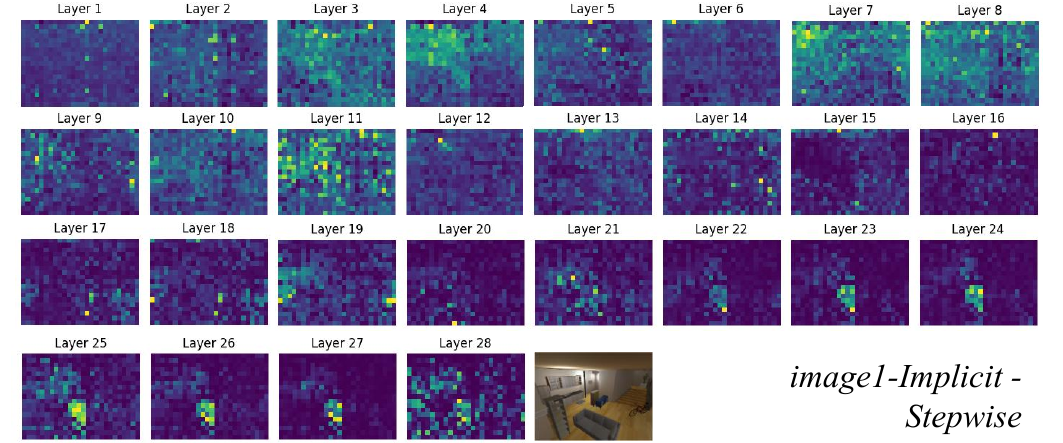}
\caption{Layer-wise attention visualization of image1 in Qwen2.5-VL-7B for the case input with the Implicit Stepwise prompt.}
\label{fig:attention_image1-Implicit-Stepwise}
\end{figure*}

\begin{figure*}[h]
\centering
\includegraphics[width=1\linewidth]{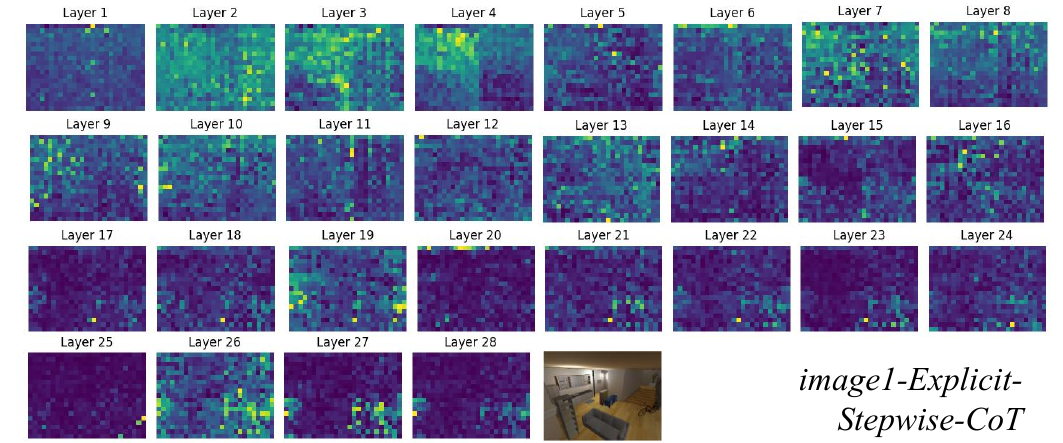}
\caption{Layer-wise attention visualization of image1 in Qwen2.5-VL-7B for the case input with the Explicit Stepwise CoT prompt.}
\label{fig:attention_image1-Explicit-Stepwise-CoT}
\end{figure*}

\begin{figure*}[h]
\centering
\includegraphics[width=1\linewidth]{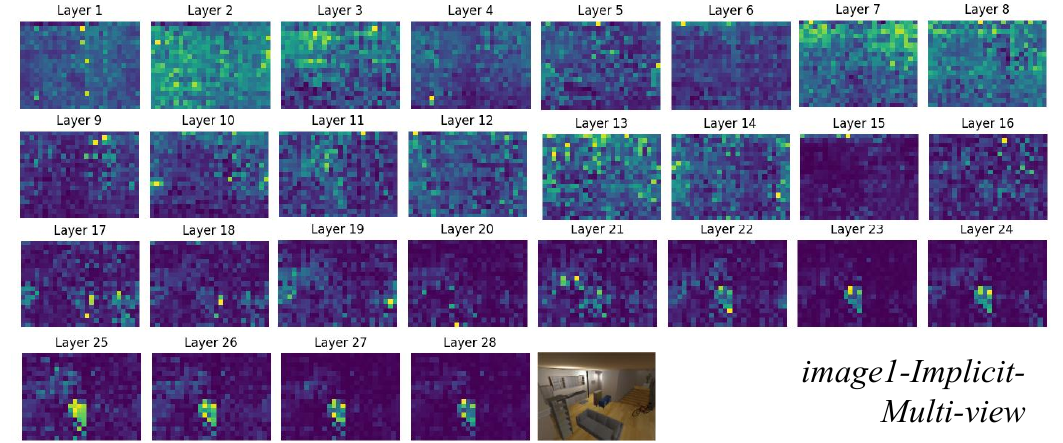}
\caption{Layer-wise attention visualization of image1 in Qwen2.5-VL-7B for the case input with the Implicit Multi-view prompt.}
\label{fig:attention_image1-Implicit-Multi-view}
\end{figure*}

\begin{figure*}[h]
\centering
\includegraphics[width=1\linewidth]{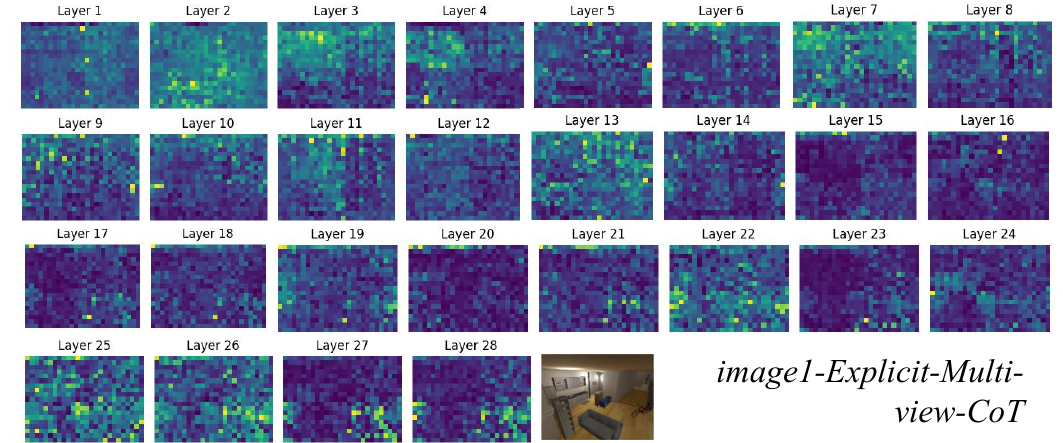}
\caption{Layer-wise attention visualization of image1 in Qwen2.5-VL-7B for the case input with the Explicit Multi-view CoT prompt.}
\label{fig:attention_image1-Explicit-Multi-view-CoT}
\end{figure*}

These detailed visualizations (Figures~\ref{fig:attention_image0-Vanilla}, ~\ref{fig:attention_image0-Implicit-Stepwise}, ~\ref{fig:attention_image0-Explicit-Stepwise-CoT}, ~\ref{fig:attention_image0-Implicit-Multi-view}, ~\ref{fig:attention_image0-Explicit-Multi-view-CoT},~\ref{fig:attention_image1-Vanilla}, ~\ref{fig:attention_image1-Implicit-Stepwise}, ~\ref{fig:attention_image1-Explicit-Stepwise-CoT}, ~\ref{fig:attention_image1-Implicit-Multi-view}, ~\ref{fig:attention_image1-Explicit-Multi-view-CoT},) offer a granular view that strongly corroborates our primary findings. As observed across the majority of layers, the attention maps for both the Vanilla and Implicit inference methods demonstrate a focused attention pattern. The model correctly concentrates on the salient objects required for spatial comparison—in this case, the sofa and the TV stand—to determine which is closer to the bike.
 
In stark contrast, the attention maps for the Explicit method are visibly more diffused throughout the layers. Crucially, the explicit reasoning process appears to misdirect the model's focus. Instead of evaluating the contextual objects, the model's attention is consistently drawn to the primary subject of the query (the bike itself). This fixation prevents the model from properly attending to and comparing the positions of the sofa and the TV stand.
 
This comprehensive layer-by-layer perspective substantiates our conclusion that while explicit reasoning chains can guide language generation, they can also disrupt the model's intrinsic visual grounding. This disruption leads to a failure to attend to critical spatial context, ultimately resulting in an incorrect answer.

%====================================================================
%====================================================================

%====================================================================
%====================================================================

%====================================================================
%====================================================================

%====================================================================
%====================================================================

%====================================================================
%====================================================================

%====================================================================
%====================================================================

%====================================================================
%====================================================================

%====================================================================
%====================================================================

%====================================================================
%====================================================================

%====================================================================
%====================================================================
\end{document}